\RequirePackage{fix-cm}
\documentclass[11pt, a4paper]{article}

\usepackage[T1]{fontenc} 
\usepackage{libertine} 

\usepackage[none]{hyphenat} 
\usepackage[a4paper, margin=1in, headheight=50pt, top=1.5in]{geometry}
\usepackage{fancyhdr}
\usepackage{setspace}
\usepackage{indentfirst}
\usepackage{caption}
\usepackage[protrusion=true]{microtype} 
\usepackage{comment}

\usepackage{lineno}

\usepackage{amsmath, amssymb, amsfonts, bm}
\usepackage{siunitx}  

\usepackage{graphicx}
\usepackage{booktabs}
\usepackage{float}
\usepackage{caption}
\usepackage{subcaption}
\usepackage{wrapfig}
\usepackage{longtable}
\usepackage{array}
\usepackage[table]{xcolor} 
\usepackage{pdflscape}
\newcolumntype{P}[1]{>{\raggedright\arraybackslash}p{#1}}

\usepackage{docmute}

\DeclareCaptionLabelFormat{bold}{\textbf{(#2)}}

\makeatletter
\newenvironment{authorsummary}{
    \small
    \begin{center}
      \bfseries Author Summary
    \end{center}
    \quotation{}
}{
    \endquotation{}
}
\makeatother

\usepackage{cite}
\newcommand{\temporalbinningrefs}{%
  komorowski2018artificial, prasad2017reinforcement, kondrup2023towards,
  wang2023optimized, nemati2016optimal, yu2019inverse, kim2021computational,
  peine2021development, sun2021personalized, eghbali2021patient,
  roggeveen2021transatlantic, yala2022optimizing, guo2022learning,
  nanayakkara2022unifying, li2022electronic, su2022establishment,
  liang2022treatment, wu2023value, denhengst2024guideline,
  kalimouttou2025optimal, desman2025distributional%
}

\newcommand{\citebinning}{\expandafter\cite\expandafter{\temporalbinningrefs}}

\usepackage{authblk} 

\usepackage[explicit]{titlesec}

\titleformat{\section}
  {\normalfont\fontsize{14}{17}\selectfont\filcenter}
  {\thesection}
  {1em}
  {\textsc{#1}}
  [\vspace{0.5em}]

\titleformat{\subsection}
  {\normalfont\fontsize{11}{13}\selectfont\bfseries}
  {\thesubsection}
  {1em}
  {\underline{#1}}
  [\vspace{0em}]

\titlespacing{\subsection}{0pt}{1.5ex plus 1ex minus 0.2ex}{0.5ex plus 0.2ex}

\titlespacing{\subsubsection}{0pt}{1.ex plus 1ex minus 0.2ex}{-0.5ex plus 0.2ex}

\usepackage{enumitem}
\setlist[enumerate]{leftmargin=5em}
\setlist[itemize]{leftmargin=3em, rightmargin=3em, nosep} 

\usepackage[hidelinks]{hyperref}

\makeatletter
\newcommand{\nexttablelabel}[1]{%
  \refstepcounter{table}%
  \label{#1}%
  \addtocounter{table}{-1}%
}
\makeatother

\title{\textbf{The impact of temporal resampling on the safety of clinical reinforcement learning: A simulated trial}}

\author[1*]{Thomas Frost}
\author[1]{Hrisheekesh Vaidya}
\author[1,2]{Steve Harris}

\affil[ ]{}
\affil[1]{University College London, London, United Kingdom}
\affil[2]{University College London Hospitals NHS Foundation Trust, London, United Kingdom}
\affil[ ]{} 
\affil[*]{Corresponding author: thomas.frost.21@ucl.ac.uk}

\date{}

\begin{document}

\maketitle
\begin{abstract}
\noindent
Reinforcement learning (RL) is a type of artificial intelligence for making optimal choices. In healthcare, researchers generally use offline RL (ORL), where models are trained and evaluated from retrospective observational data. To accommodate naturally irregular clinical records, researchers often resample the data into uniform time intervals before training (known as binning). However, binning may create counterfactual trajectories that do not represent clinical practice. We first quantified these distortions in patients receiving continuous insulin infusions in the MIMIC-IV intensive care database. Binning reversed the apparent causal order for over 70\% of paired glucose measurements and insulin adjustments within the same interval, and frequently masked significant dysglycaemic events and merged distinct treatment decisions together. We then conducted an \emph{in silico} clinical trial using 30 virtual patients with type 1 diabetes from the FDA-approved UVA/Padova simulator. The simulator was modified to include stochastic intervals between decisions and used to generate a training dataset for offline RL\@. We trained three ORL algorithms on both the unprocessed dataset and equivalent datasets resampled at 10-minute, 2-hour, and 4-hour intervals. When deployed back into the simulated environment, temporal resampling was found to reduce model performance by up to 60\% relative to unprocessed data, with 4-hour binning causing all agents to perform worse than the dataset's baseline. Retrospective evaluation on resampled data actively obscured this effect, predicting 1.5--3x better returns than agents achieved in practice. We recommend that future research in this area prioritises datasets with natural clinical timings between decisions, which may be a necessary step before these models can be safely deployed into patient care.
\end{abstract}

\begin{authorsummary}
\noindent
Artificial Intelligence (AI) models could help future doctors by recommending personalised treatments for patients. These models are currently trained and tested using historical patient records rather than real patients. Medical decisions happen at unpredictable intervals, but before training an AI model, researchers usually shift and align the timings of recorded decisions into a regular schedule (`temporal resampling'). This makes training the AI models easier with standard techniques. We investigated whether this practice leads to lower quality models that would perform poorly in the real clinical world. We ran a virtual clinical trial using an FDA-approved diabetes simulator to test how well models that had been trained on resampled data perform in environments with unpredictable decision intervals. We found that models trained on artificially regular decisions perform worse than expected when trialled in environments with these irregular timings. Concerningly, standard tests used to predict model safety before clinical use do not detect this, producing falsely optimistic scores. We then assessed models trained on natural, unedited data. These models performed much better and the safety tests accurately predicted their true performance. We therefore recommend training and evaluating AI models using unedited data to preserve the authentic timing of clinical practice.
\end{authorsummary}

\vspace{2em}

\section{Introduction}

Reinforcement learning (RL), the branch of artificial intelligence behind recent landmark achievements in games~\cite{shakya2023reinforcement}, robotics~\cite{tang2025deep}, and autonomous driving~\cite{hu2025survey}, learns to maximise the long-term rewards of sequential decisions using trial-and-error~\cite{sutton2018reinforcement}. In most domains, the agent can interact directly with its environment, known as online RL\@. In clinical medicine, this is impractical: allowing the agent to explore potentially harmful actions to find the best treatment would expose patients to risks unjustifiable outside of a formal trial framework~\cite{jayaraman2024primer}. Researchers therefore depend on \textit{offline} RL (ORL), in which agents infer better treatments entirely from retrospective clinical datasets~\cite{levine2020offline}. ORL has been applied to a wide range of clinical problems, including fluid dosing in sepsis~\cite{komorowski2018artificial, liu2021offline, fatemi2021medical}, chemotherapy personalisation~\cite{liu2017deep, shiranthika2022supervised}, ventilation management~\cite{prasad2017reinforcement, kondrup2023towards}, and insulin delivery in diabetes~\cite{wang2023optimized}.\label{reforms1c} However, despite intentions to develop these models for patient care, there is very little prospective evidence of their real-world performance~\cite{otten2024reinforcement, jayaraman2024primer}. In a review of 50 healthcare ORL studies published between 2016 and 2026, we found that only three had evaluated their models in a prospective human trial (Supplementary Table~\ref{tab:lit-review})~\cite{gao2023offline, wang2023optimized, fan2026reinforcement}; two had used a simulated environment~\cite{zhu2023offline, beolet2024end}; and the remaining 45 relied solely on retrospective evaluation using historical data. This persistent gap between research publication and clinical deployment is unsurprising given the challenges of conducting real-world clinical trials---but as a result, the true efficacy of these models remains unclear.

A particularly prevalent feature of healthcare ORL pipelines is temporal resampling: the aggregation or interpolation of electronic health record (EHR) data into standardised time intervals prior to model training, often known as ``binning''~\cite{lipton2016modeling, shukla2020survey}. Of the 50 studies reviewed, 43 used datasets that had been temporally resampled. This is in spite of widely available architectures for processing irregularly sampled data directly~\cite{neil2016phased, che2018recurrent, kidger2020neural, shukla2021multi, tipirneni2022self}. The popularity of discretising data likely relates to the convenience offered by regular decision labels, which greatly simplifies the implementation of the Markov decision process (MDP) framework underpinning most ORL algorithms~\cite{bellman1957markovian, puterman2014markov, sutton2018reinforcement, jayaraman2024primer}. Although alternative RL frameworks for variable decision intervals exist~\cite{sutton1999between}, the use of these frameworks in healthcare is exceedingly rare. Of the seven studies in our review that did not resample data, three used data that were already regular~\cite{shiranthika2022supervised, gao2023offline, zhu2023offline}; three treated irregularly spaced visits as uniform time steps~\cite{job2024optimal, zhou2025optimizing, ghasemi2025personalized}; and only one explicitly modelled for irregular decision timing~\cite{petch2024optimizing}.

Unfortunately, this simplification may come at a cost. Clinical interactions are naturally irregular: a clinician may act within minutes of a deterioration or withhold intervention for hours during periods of stability. Additionally, the practicalities of providing clinical care will lead to further random variation in the timings of clinical events, measurements, and actions. By resampling data into fixed intervals, these nuances are completely hidden from the model, which is instead presented with a greatly simplified observation space and action/temporal dynamics. Concerns have already been raised that discretisation may bias the policies that RL models learn~\cite{schulam2018discretizing, jeter2019artificial}, with several studies showing that the choice of resampling interval materially affects both the policy identified and its estimated performance during retrospective evaluation~\cite{lu2020deep, wu2023value, sun2025exploring, tang2026off}.

In this paper, we argue that temporal resampling violates a core assumption of ORL\@: namely, that training data should faithfully represent the intended deployment environment~\cite{levine2020offline}. By presenting models with a fictitious, regularly-timed version of reality, temporal resampling may lead to models performing suboptimally when deployed into settings with unpredictable decision intervals---and, critically, this failure may be undetectable if retrospective evaluation is conducted on the same resampled data. This creates a significant and avoidable patient safety risk, as a model could progress to human trials on the basis of misleading retrospective evidence.\label{reforms1a}\label{reforms1b}

To test this, we first quantified how binning corrupts training data by looking at real insulin titration decisions in intensive care unit (ICU) patients from the Medical Information Mart for Intensive Care (MIMIC-IV) database~\cite{johnson2023mimic}. We derived a cohort of real patients receiving continuous insulin infusions and demonstrate that resampling into fixed bins can mask pathological events and lead to inversions of cause (e.g., low/high glucose) and effect (insulin adjustment).

We then generated a dataset using 30 virtual patients with type 1 diabetes from the FDA-approved UVA/Padova simulator~\cite{man2014uva}, which we modified to have randomly irregular decision intervals. By using a virtual environment, we were able to conduct an \textit{in silico} clinical trial investigating the effects of dataset resampling on ORL model performance. We focus on model-free algorithms, which feature in the majority of papers found in Supplementary Table~\ref{tab:lit-review}. Three ORL algorithms were trained on resampled and unprocessed versions of the same dataset, and evaluated by their interaction with the true, irregularly-timed clinical environment. Finally, we compared these prospective results against those predicted by standard retrospective evaluation conducted on the training dataset. We conclude our paper with recommendations for future healthcare ORL research, particularly where progression to clinical trials is intended.\label{reforms3c}

\section{Results}
\subsection{Temporal binning leads to counterfactual data in a real-world electronic healthcare dataset}

We temporally resampled a cohort of real ICU patients receiving continuous insulin infusions at 1-, 2-, 3-, and 4-hour intervals, in order to examine the effects of binning on the fidelity of the training data. Recent research has shown that aggregating observations and actions over the same window carries a significant risk of look-ahead bias, as any observations recorded after the commencement of an action(s) will be confounded by the action itself~\cite{tang2026off}. To address this, the recommended approach is to aggregate observations over the window immediately preceding the action window.

\begin{figure}[!t]
    \centering
    \includegraphics[width=\textwidth]{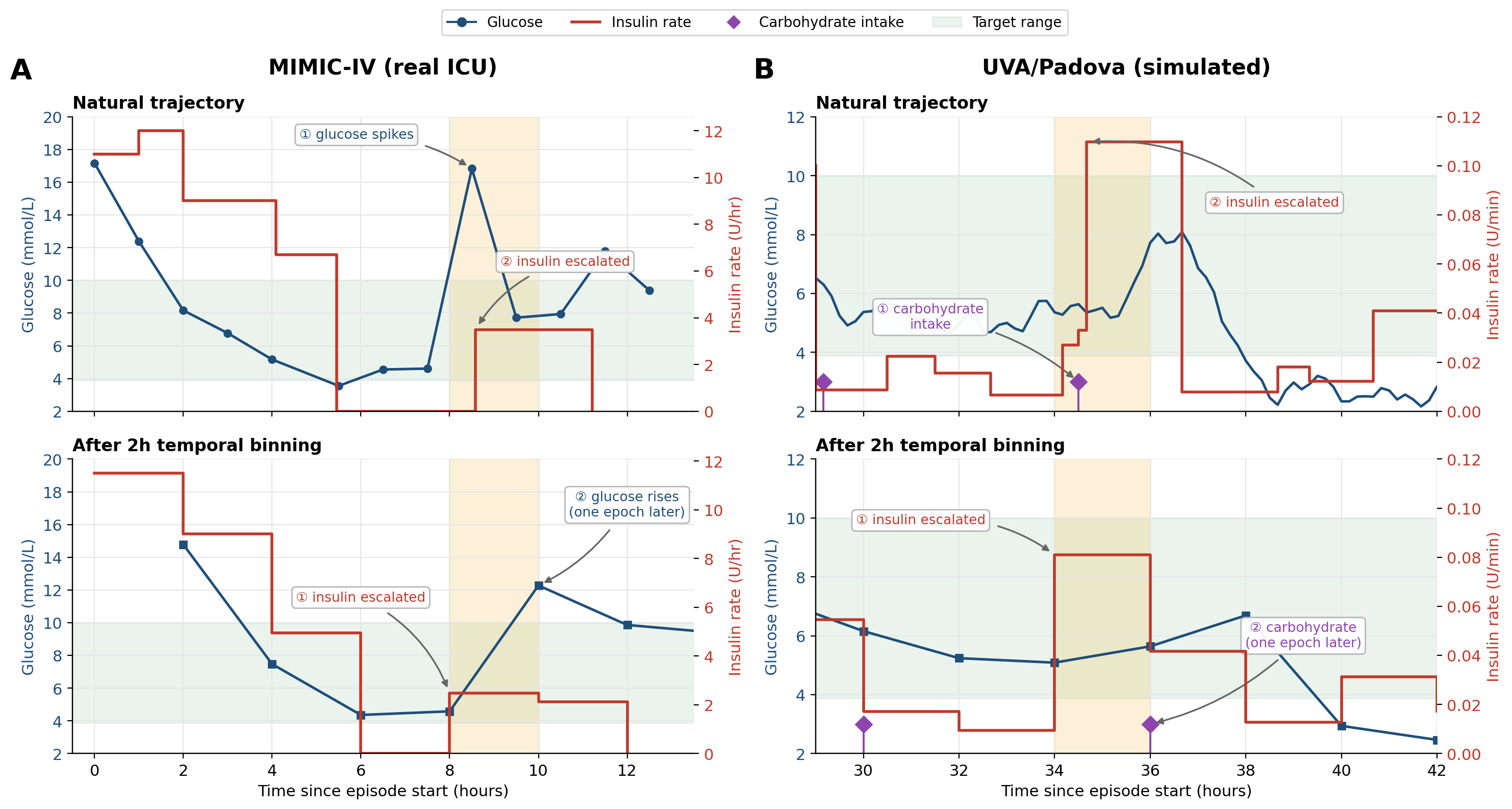}
    \caption{\textbf{Example patient trajectories from MIMIC-IV and the UVA/Padova simulator.}
        A) The top and bottom panels display a patient trajectory from MIMIC-IV receiving an insulin infusion, before and after temporal binning (respectively). A causal inversion artefact can be seen, where insulin was escalated in response to a spike in the blood glucose level. In the binned trajectory, this action appears to occur in response to glucose levels from the previous time window, with the glucose spike occurring after the action. B) A patient trajectory from the UVA/Padova simulator, also before and after binning. A causal inversion artefact is observed, this time in response to a carbohydrate intake.}\label{fig:combined_example_trajectory}
\end{figure}

\begin{figure}[htbp]
    \centering
    \includegraphics[width=\textwidth]{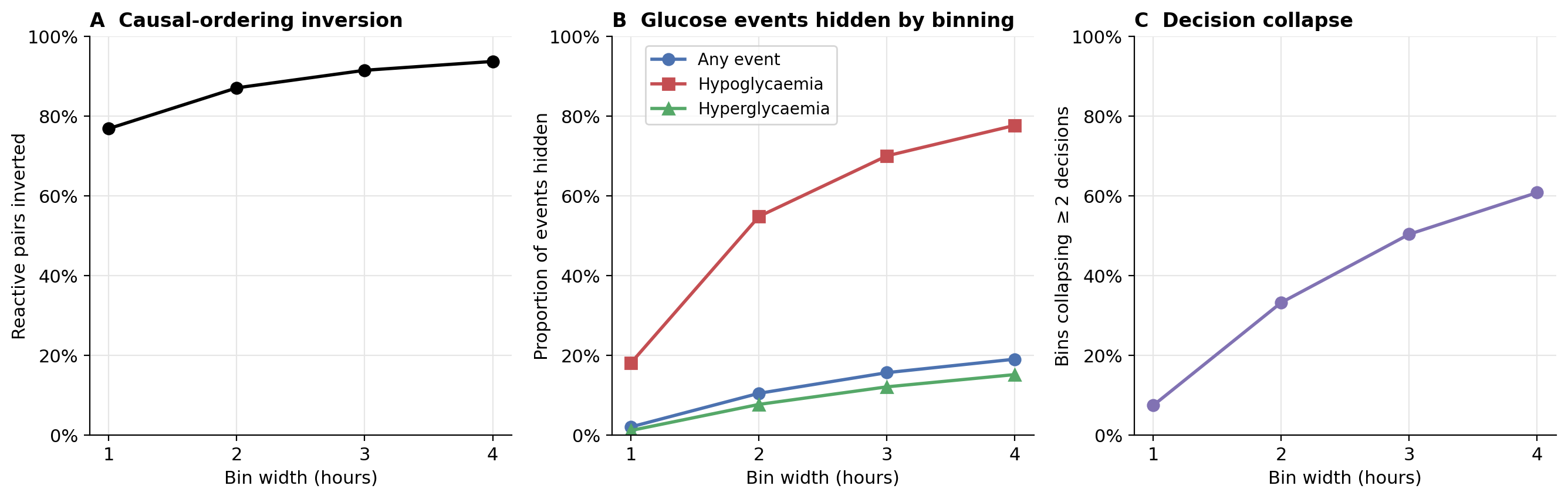}
    \caption{\textbf{Quantification of artefacts in MIMIC-IV following temporal binning.}
        A) The proportion of causally inverted artefacts (from reactive glucose-insulin pairs that occupy the same time window) is shown to be high even with 1-hour binning, and increases monotonically with greater bin widths. B) The proportion of hypo- and hyperglycaemic events that are masked as a result of averaging blood glucose levels within each bin window. C) The proportion of bins containing 2 or more insulin adjustment decisions (which are then aggregated into a single action).}\label{fig:mimic_binning_metrics}
\end{figure}

For our cohort, we identified all cases where a glucose measurement (an observation) was followed by an insulin rate adjustment (an action). If these reactive pairs occupy the same temporal window, they may become causally inverted: post-aggregation, the action is viewed in response to the observation(s) of the \textit{previous} time window, and the precipitating observation is incorrectly treated as an outcome of the action rather than its trigger. This is visualised in Fig~\ref{fig:combined_example_trajectory}A. We found that the inversion of causal ordering is very common (>70\%) and becomes more prevalent as the width of the time bin is increased (Fig~\ref{fig:mimic_binning_metrics}A). Additionally, averaging multiple recorded events within a single window masks a significant number of episodes of both hypoglycaemia and hyperglycaemia (Fig~\ref{fig:mimic_binning_metrics}B), as well as blurring together multiple insulin adjustment decisions into a single perceived action (Fig~\ref{fig:mimic_binning_metrics}C).

\subsection{Resampling degrades the online performance of offline reinforcement learning models}

To investigate the effects of resampling on prospective performance of offline reinforcement learning models, we conducted a set of \textit{in silico} experiments using 30 virtual patients from the UVA/Padova Type 1 Diabetes Simulator~\cite{man2014uva}, an FDA-approved clinical simulator for type 1 diabetes mellitus (T1DM). This simulator is accepted as a substitute for pre-clinical animal research studies, including closed-loop ``artificial pancreas'' algorithms, prior to human clinical trials. The simulator can be used to generate datasets for training as well as a safe environment in which to evaluate models prospectively. Preliminary validation of the methodology was performed using LavaGap~\cite{chevalier2023minigrid}, a partially observable gridworld environment requiring navigation around hazards (see Supplementary Fig~\ref{fig:lavagap_example}).\label{reforms3g}

Environments were implemented in two configurations: a regular version with fixed decision intervals and an irregular version with variable intervals sampled from a uniform random distribution. Within each irregular environment, we pretrained an online RL model using proximal policy optimisation (PPO)~\cite{schulman2017proximal}. This agent acts as a simulated proxy for a competent clinician and is used to generate an unprocessed training dataset with variable intervals between decisions. The unprocessed dataset was then cloned and preprocessed into two variants: (1) a binned dataset, which downsamples observations and rewards via aggregation into broad, fixed time windows; and (2) an interpolated dataset, which upsamples the decision frequency such that a decision is recorded at every environmental time step. Full specifications and data generation details are provided in the Methods.

An example of a simulated patient trajectory (before and after temporal binning) is shown in Fig~\ref{fig:combined_example_trajectory}B. The top panel displays real-time changes in blood glucose and insulin sampled at the environment's 10-minute base frequency, with variable-length plateaus in the insulin infusion rate reflecting the irregular decision intervals. The bottom panel shows the effect of a typical 2-hour binning regimen. As with MIMIC-IV, the simulated dataset suffers similar effects of causal inversion and deceptive smoothing, with observations and actions flattened into single, ambiguous data points (Supplementary Fig~\ref{fig:simglucose_binning_metrics}).

We then assessed the impact of temporal resampling (both binning and interpolation) on several offline reinforcement learning algorithms: behavioural cloning (BC), implicit Q-learning (IQL), and conservative Q-learning (CQL). IQL and CQL are widely regarded as state-of-the-art model-free methods for offline RL, demonstrating robust performance across diverse discrete and continuous control tasks. BC serves as a supervised learning baseline by attempting to directly mimic the dataset's behaviour rather than optimise for long-term rewards. For each algorithm, identical neural networks were trained on the three dataset variants and evaluated in both the regularly and irregularly timed versions of each environment. We used a semi-Markov decision process (SMDP) framework for all training~\cite{jewell1963markov, bradtke1994reinforcement, sutton1999between}; this is functionally equivalent to a standard MDP under the setting of fixed interstate intervals (i.e., resampled data).

\begin{figure}[htbp]
    \centering
    \includegraphics[width=\textwidth]{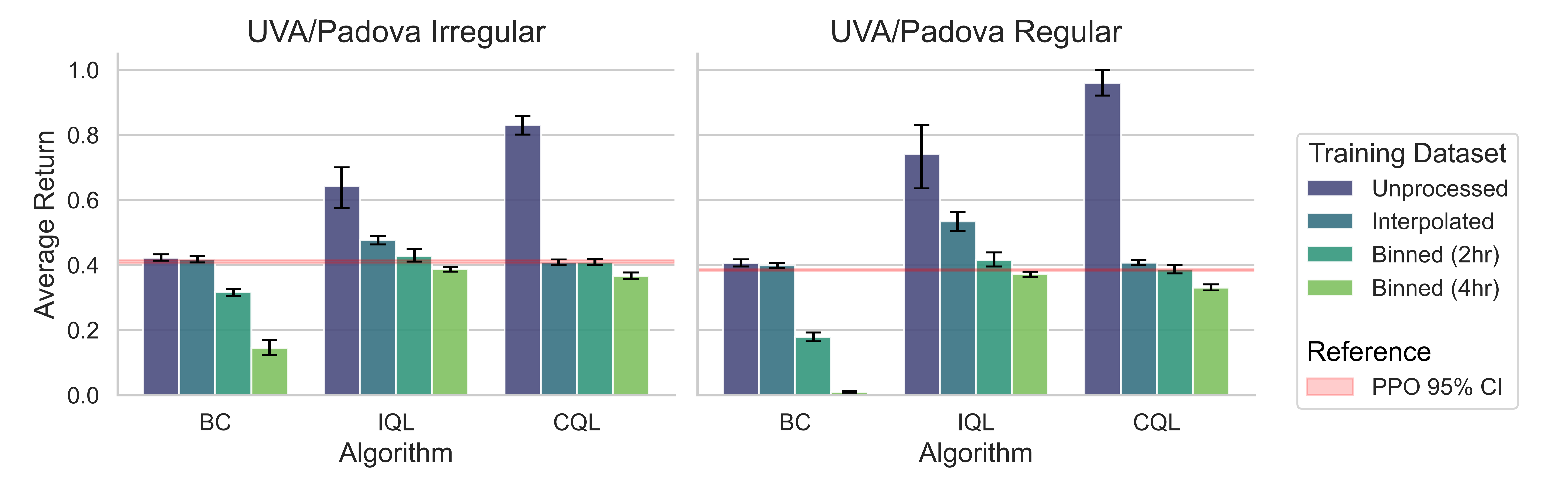}
    \caption{\textbf{Impact of temporal resampling on offline RL performance.}
        Models trained via behavioural cloning (BC), implicit Q-learning (IQL), or conservative Q-learning (CQL) were evaluated on the UVA/Padova insulin control task. Models were trained using unprocessed, interpolated, or temporally binned datasets and deployed in both regularly and irregularly timed versions of the environment. Models trained on the unprocessed dataset consistently achieved the highest returns, which were heavily degraded using binned or interpolated datasets. The pink band indicates the expert proximal policy optimisation (PPO) baseline used to generate the training data. Average returns are normalised (0.0 for a random policy; 1.0 for the highest observed score). Shaded regions and error bars represent 95\% confidence intervals (CIs).}\label{fig:padova_results}
\end{figure}

The results for the clinical UVA/Padova environment are shown in Fig~\ref{fig:padova_results}, compared against the simulated clinician baseline. The unprocessed dataset consistently produced the highest average returns for both CQL and IQL, substantially outperforming the clinical baseline. Performance declined monotonically as the degree of temporal abstraction was increased: interpolation led to better results than 2-hour binning, which in turn outperformed 4-hour binning (an interval commonly used in ORL studies). We found that all models trained on the 4-hour binned dataset consistently performed worse than the baseline performance of the dataset, and up to 60\% worse than the models trained on unprocessed data. Overall, no models trained on resampled data were able to improve on the baseline performance of their training dataset, with the exception of modest improvements using the IQL algorithm on interpolated data. We obtained very similar results in the gridworld navigation task (Supplementary Fig~\ref{fig:lavagap_results}).

For UVA/Padova, our reward score was calculated from clinically important outcomes (the combination of the Magni risk index with penalties for extreme hypo/hyperglycaemia, as detailed in Methods). Improvements in this score therefore correspond to direct improvements in either the proportion of time in optimal blood glucose range (pTIR; 70--180 mg/dL or 3.9--10 mmol/L) and/or the proportion of episodes ending prematurely due to extreme hypo/hyperglycaemia (pET; <10 or >600 mg/dL, or <0.6 or >33.3 mmol/L). These additional clinical endpoints are shown in Supplementary Fig~\ref{fig:time_in_target_range} and Supplementary Fig~\ref{fig:proportion_early_termination}, which show that only models trained on unprocessed data have consistent improvements in pTIR \textit{and} pET.

\subsection{Retrospective evaluation can be overoptimistic on binned data}

\begin{figure}[!t]
    \centering
    \includegraphics[width=0.8\linewidth]{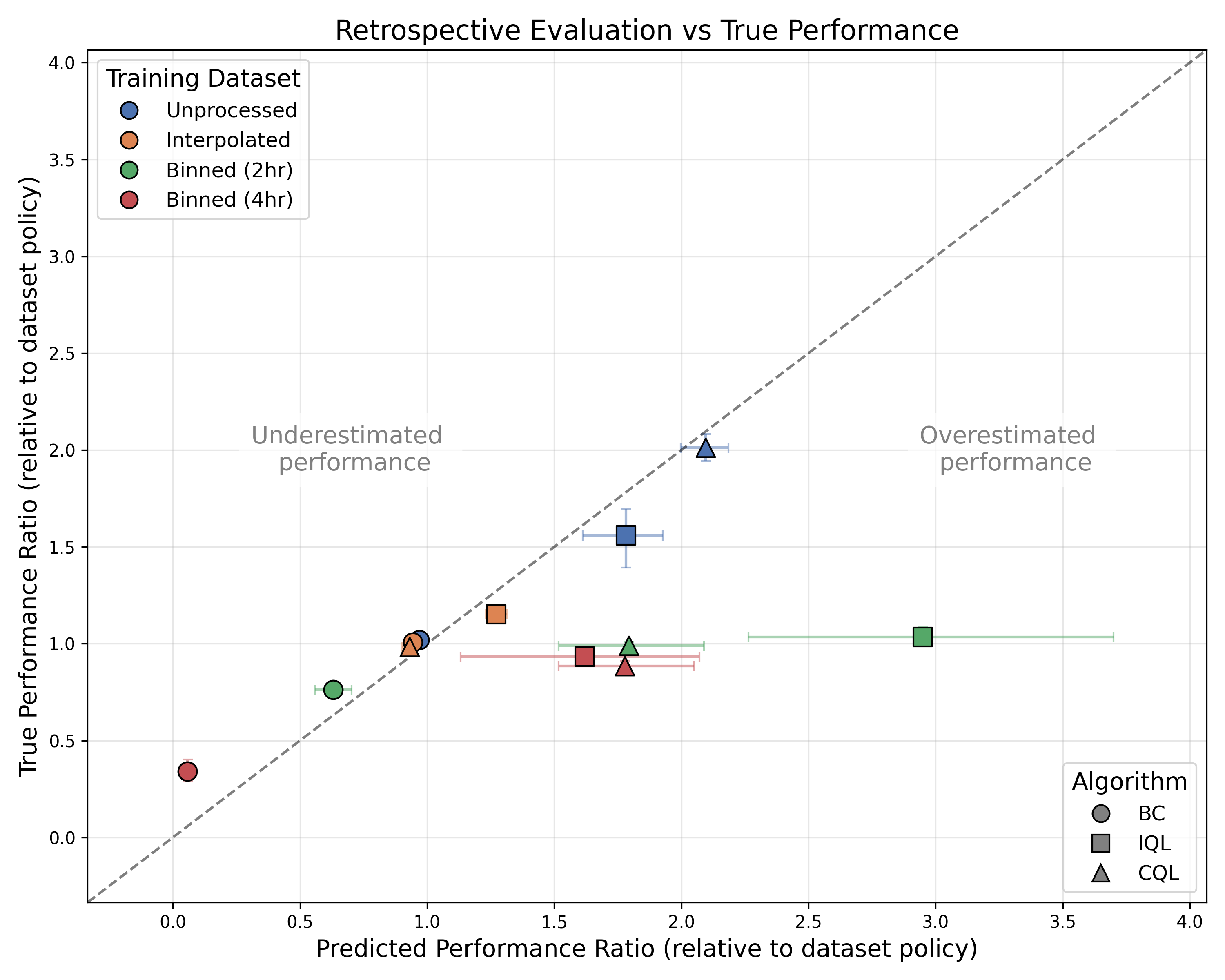}
    \caption{\textbf{Calibration plot showing reliability of off-policy evaluation across different types of dataset preprocessing.}
        The plot compares the true online performance of trained agents in the UVA/Padova environment against the performance predicted by fitted Q-evaluation (FQE)\@. Performance is normalised such that 0.0 represents a random policy and 1.0 represents the dataset's behaviour policy. While agents trained on unprocessed (blue) and interpolated (orange) data show high calibration (clustering near the diagonal), those trained on temporally binned data (green, red) exhibit severe overestimation bias. Error bars represent 95\% confidence intervals.}\label{fig:fqe_results}
\end{figure}

Next, we examined how the accuracy of retrospective (or ``off-policy'') evaluation varies in response to temporal resampling. We employed fitted Q-evaluation (FQE)~\cite{le2019batch} to retrospectively estimate policy performance. This well-established technique involves training a model to predict our agent's expected returns within the constraints of a static dataset. For each algorithm-dataset pair, we trained multiple FQE models to predict that agent's expected return and compared the predictions against the agent's true performance in the irregular UVA/Padova environment. Fig~\ref{fig:fqe_results} presents these results as a calibration plot relative to the baseline performance for each dataset. Overall, FQE models trained on the unprocessed and interpolated datasets showed reasonably high calibration, accurately estimating the returns of the IQL and CQL agents. In contrast, FQE severely overestimated the performance of agents trained on binned data, predicting 1.5--3x improvements for agents that, in reality, only matched or underperformed their respective baselines.

This systematic overestimation likely reflects the fact that IQL and CQL are solving for actions that appear highly favourable within the binned datasets, even if such actions are suboptimal in reality. When FQE is conducted on the same post-processed data, its predictions become similarly optimistic. However, because the binned dataset contains trajectories that are counterfactual, these predictions amount to hallucinations rather than accurate estimates.

\section{Discussion}
This study demonstrates that temporal resampling, a widespread preprocessing step in healthcare-based ORL, introduces avoidable harm to model safety and efficacy. By using training data from a virtual clinical environment, we evaluated how resampling affects prospective performance within a risk-free simulation. Our results show that 1) models trained on resampled data show reduced performance in natural settings with unpredictable decision timings, and 2) standard retrospective evaluation actively obscures this effect with overoptimistic estimates for model performance. The latter point is most concerning as such evaluations could be used to justify clinical trials for suboptimal models. We will now propose several overlapping mechanisms to explain these observed effects.

First, we found that temporal binning can produce a significant number of counterfactual artefacts. This was observed in both our simulated dataset and in real EHR data gathered from MIMIC-IV. For example, binning over 4 hours---a standard practice in sepsis-based ORL research~\cite{raghu2017continuous, komorowski2018artificial, peng2018improving, tang2020clinician, killian2020empirical, roggeveen2021transatlantic, liu2021combining, fatemi2021medical, satija2021multi, liang2023treatment, tu2025offline, fang2025offline}---can mask severe hypo-/hyperglycaemia by averaging these values into a stable mean~\cite{jeter2019artificial}. Aggregating observations over the preceding window from actions has the benefit of avoiding look-ahead bias~\cite{tang2026off}, but risks producing causally inverted data. The creation of these spurious trajectories result in the agent learning harmful out-of-distribution behaviours, which appear beneficial in off-policy evaluation on the same dataset, but are revealed to be harmful during prospective evaluation.

Non-aggregated interpolated data does not suffer from the same counterfactual artefacts, and as a result, FQE estimates of performance derived from these data are relatively accurate. However, models trained on such data will still learn to expect deceptively frequent opportunities to adjust their actions. If this expectation is violated at deployment, the agent may produce harmful behaviour by taking actions that persist far longer than intended. Perhaps surprisingly, models trained on the interpolated dataset and deployed into an environment with similar high-frequency intervals also performed worse than expected. We suspect that this is due to an increased sensitivity to ``distributional shift''~\cite{kumar2019stabilizing}: if the model makes errors that expose it to unfamiliar states not present in the training dataset, it is likely to make further compounding mistakes with eventual performance collapse. If a single sustained action (a 2-hour infusion) is reframed as a long series of independent actions (twelve 10-minute decisions), the cumulative opportunity for such errors increases proportionally. Our results suggest that this increased risk outweighs any benefit from having more opportunities to intervene.

As an alternative solution to temporal resampling and its associated risks, we propose retaining the natural intervals between decision labels and utilising reinforcement learning frameworks that explicitly account for them (such as the SMDP framework employed in this work). As demonstrated by our results, this approach provides the model with the most accurate representation of the environment's temporal dynamics, resulting in the greatest capacity to improve upon the dataset's baseline behaviour, whilst critically retaining accuracy of retrospective evaluation.

This study has several limitations. Firstly, our simulated environments may have increased sensitivity to information loss from binning due to their simpler observation space, and because our decision intervals are randomly sampled from a state-independent uniform distribution. This may limit the generalisability of our results to more complex real-world clinical settings. Additionally, we only use model-free ORL algorithms and do not include any model-based techniques. Model-free techniques constitute the significant majority of the existing healthcare ORL literature (Supplementary Table~\ref{tab:lit-review}), with numerous benchmark methods (e.g., IQL/CQL). In contrast, all model-based approaches identified in our review were bespoke rather than a standardised set of techniques, rendering them less suitable for benchmark comparisons. Finally, our simulated experiments focus on the impact of temporal resampling on irregularly sampled \textit{decision-making} only; for irregularly sampled observations, we direct the interested reader to several modern architectures designed to process such inputs without temporal resampling~\cite{che2018recurrent, kidger2020neural, shukla2021multi, tipirneni2022self}.
\label{reforms8b}

\section{Conclusion}
Our findings confirm that common approaches to temporal resampling can lead to a fundamental misalignment between the training data and the intended deployment environment. By forcing natural variations in decision intervals into a discretised schedule, researchers risk training models that are only optimal for those schedules---and adapt poorly to any unpredictability. Critically, the fragility of these models is undetectable by standard retrospective evaluation methods when performed on similarly resampled data. Any falsely optimistic evaluations could result in the premature justification of a clinical trial for these models. We therefore recommend that future research move beyond default binning practices and prioritise methods that preserve these natural decision intervals. In doing so, we can improve the safety and viability of these advanced models as they transition from digital modelling to the patient bedside.

\section{Materials and Methods}
\subsection{Preliminaries}
Reinforcement learning conventionally uses a Markov decision process (MDP) to model the interaction between the agent and the environment~\cite{bellman1957markovian, puterman2014markov, sutton2018reinforcement}. We define an MDP by the tuple $\mathcal{M} = (\mathcal{S}, \mathcal{A}, T, \mu_0, R, \gamma)$, where:

\begin{itemize}
    \item $s_t \in \mathcal{S}$ is the state at time $t$ (e.g., the physiological state of the patient) within the state space $\mathcal{S}$;
    \item $a_t \in \mathcal{A}$ is the action taken at time $t$ from the available action space $\mathcal{A}$;
    \item $\mu_0$ is a probability distribution over $\mathcal{S}$, with the initial state $s_0 \sim \mu_0$;
    \item $T(s_{t+1}|s_t,a_t): \mathcal{S} \times \mathcal{A} \rightarrow \Delta(\mathcal{S})$ is the transition kernel, representing the probability of reaching state $s_{t+1}$ given the current $s_t$ and $a_t$;
    \item $R: \mathcal{S} \times \mathcal{A} \times \mathcal{S} \rightarrow \mathbb{R}$ is the reward function, where $r_{t+1} = R(s_t, a_t, s_{t+1})$;
    \item $\gamma \in \left[0, 1\right)$ is the discount factor, determining the present value of future rewards.
\end{itemize}

If the observations from the environment are only a partial reflection of the true underlying state (such as in healthcare), this is treated as a partially observable MDP (POMDP), with additional elements:

\begin{itemize}
    \item $o_t \in \mathcal{O}$, the observations perceived at time $t$ (e.g., blood pressure, heart rate) from the observation space $\mathcal{O}$;
    \item $\mathcal{Z}(o_t | s_t, a_{t-1}) : \mathcal{S} \times \mathcal{A} \rightarrow \Delta(\mathcal{O})$, the observation model representing the probability of perceiving $o_t$ given the current true state $s_t$ and previous action $a_{t-1}$.
\end{itemize}

The policy $\pi$ of the agent generates actions $a_t$ in response to states $s_t$ (or observations $o_t$), aiming to maximise the expected cumulative discounted return:
\begin{equation}
    J(\pi) = \mathbb{E}_{\tau \sim (\pi, T, \mu_0)} \left[ \sum_{t=0}^\infty \gamma^t R(s_t,a_t,s_{t+1}) \right] \label{eq:mdp-objective}
\end{equation}
where the trajectory $\tau = (s_0, a_0, s_1, a_1, \dots)$ represents some sequence of environmental interactions.

If the time intervals between agent decisions are variable, the environment can be considered as a semi-Markov decision process (SMDP)~\cite{jewell1963markov, bradtke1994reinforcement, sutton1999between}. In this setting, we distinguish between time steps $t$ (the underlying clock time) and decision epochs $k$. If the $k$-th decision occurs at time $t_k$, the system remains in a state $s_k$ for some finite duration $\Delta t_k = t_{k+1} - t_k$ after taking action $a_k$, before transitioning to the next state $s_{k+1}$ at the next decision epoch. The transition dynamics are defined by a joint distribution over the next state and the duration:
\[ T(s_{k+1}, \Delta t_k \mid s_k, a_k) \]

The reward accumulated during this interval, $r_{k+1}$, is the discounted sum of atomic rewards $p$ collected at each underlying time step between $t_k$ and $t_{k+1}$:
\begin{equation} \label{eq:smdp-reward-accrued}
    r_{k+1} = R(s_k, a_k, s_{k+1}, \Delta t_k) = \mathbb{E} \left[ \sum_{i=0}^{\Delta t_k - 1} \gamma^i p_{t_k+i+1} \mid s_k, a_k, s_{k+1}, \Delta t_k \right]
\end{equation}

The agent's objective in this setting is to maximise the expected return discounted by the cumulative real-time duration. Letting $t_k$ denote the actual time at the $k$-th decision epoch (where $t_k = \sum_{i=0}^{k-1} \Delta t_i$), the objective becomes:
\begin{equation}
    J(\pi) = \mathbb{E}_{\tau \sim (\pi, T, \mu_0)} \left[ \sum_{k=0}^\infty \gamma^{t_k} R(s_k, a_k, s_{k+1}, \Delta t_k) \right] \label{eq:semi-mdp-objective}
\end{equation}

To avoid ambiguity, within this manuscript we use \textit{time steps} to denote the smallest unit of environmental transition; \textit{actions} to denote the inputs applied to the environment at each time step; and \textit{decisions} to denote the agent's chosen action, which may be repeated across multiple time steps in an SMDP\@.

\subsection{Simulated environments}\label{reforms5ai}
We conducted our experiments in two simulated environments, modified to support irregular decision intervals.

\subsubsection{LavaGap}
LavaGap~\cite{chevalier2023minigrid} is a partially observable, two-dimensional gridworld in which the agent must reach a green goal square while avoiding orange lava pits (Supplementary Fig~\ref{fig:lavagap_example}). The agent views the observable pixels and selects one of several discrete actions: move forward, rotate left or right, or do nothing. Each episode returns a reward of 1 if the agent reaches the goal, and 0 if it falls into lava or exceeds the 100-step limit.

\subsubsection{UVA/Padova}
The UVA/Padova simulator~\cite{man2014uva} is an FDA-approved clinical diabetes simulator comprising 300 virtual patients with type 1 diabetes mellitus (T1DM)\@. For our experiments, we customised a previously published implementation that describes 30 patients (10 adults, 10 adolescents, 10 children)~\cite{xie2018simglucose}. The 30 patients were partitioned into three distinct cohorts: a training cohort of 18 patients (used for all training experiments), and validation and testing cohorts of 6 patients each. All cohorts maintained an equal split of adult, adolescent, and child patients.\label{reforms4a}

Each patient has a number of individualised physiological parameters, such as glycogen stores, insulin sensitivity, and gastric emptying rate. The observation space consists of current blood glucose, the insulin infusion rate, time of day, and any carbohydrate intake since the previous observation. The continuous action space represents the current basal insulin infusion rate. Each environmental time step lasts 10 minutes. Episodes terminate after 48 hours, or earlier if glucose levels become dangerously low (<10 mg/dL or 0.6 mmol/L) or high (>600 mg/dL or 33.3 mmol/L).\label{reforms4b}\label{reforms6b}

The reward function is derived from the Magni risk index~\cite{kovatchev1997symmetrization} (Eq.~\ref{eq:magni-risk-index}), which is a function of the patient's blood glucose level (in mg/dL). We shift this by a constant to set positive rewards in the euglycaemic range (70--180 mg/dL or 3.9--10 mmol/L), and apply a max operator to increase the magnitude of positive rewards relative to negative rewards (Eq.~\ref{eq:reward-function}). The reward landscape is visualised in Supplementary Fig~\ref{fig:simglucose_example_extended}. The environment additionally returns a penalty of --10,000 for terminating early due to extreme blood glucose levels.

\begin{equation}
    g(x) = 15.09 \cdot {(\ln{(x)}^{1.084} - 5.381)}^2 \label{eq:magni-risk-index}
\end{equation}

\begin{equation}
    R(x) = \max(5.1 - g(x), 10 \cdot (5.1 - g(x))) \label{eq:reward-function}
\end{equation}

\subsubsection{Regular and Irregular Variants}
In both environments, we implemented a regular version, in which each agent decision covers exactly one time step, and an irregular version, in which decisions are applied after a variable number of elapsed time steps. In LavaGap Irregular, intervals between decisions alternate stochastically between 1 and 3 time steps. For UVA/Padova Irregular, we chose to sample from a uniform distribution (over 1 to 12 time steps)---although not as clinically realistic, this is the only distribution guaranteed to provide a fixed amount of uncertainty to each model independent of the size of resampling interval (compared to any parameterised distribution e.g., Gaussian).

\subsection{Methods for dataset generation}\label{reforms2a}\label{reforms3a}\label{reforms3b}

\subsubsection{Real EHR data from MIMIC-IV}

To quantify the direct impact of temporal binning on EHR data, we analysed real-world data from the Medical Information Mart for Intensive Care (MIMIC-IV) database~\cite{johnson2023mimic}, hosted on the PhysioNet repository of healthcare-related datasets~\cite{johnson2024mimic, goldberger2000physiobank}. MIMIC-IV consists of de-identified healthcare data collected from the Beth Israel Deaconess Medical Centre in Massachusetts, USA. We extracted a cohort of 12,238 patients receiving continuous intravenous insulin infusions in the intensive care unit, alongside irregularly sampled blood glucose levels. Patient records were segmented into discrete episodes based on activity gaps exceeding 24 hours, with the temporal origin anchored to the first event of each episode. This yielded 19,449 distinct episodes. We applied 1-hour, 2-hour, 3-hour, and 4-hour temporal binning and measured three data-corruption artefacts: causal ordering inversion, masking of glycaemic excursions, and decision distortion. Continuous insulin infusions were treated as step signals and interpolated using a carry-forward approach. Observations were aggregated as the mean of the temporal window immediately preceding the corresponding aggregated action~\cite{tang2026off}, while binned actions were calculated as the time-weighted mean of the interpolated signal over the current window.

Reactive pairs were strictly defined by clinical thresholds: hyperglycaemia (>10.0 mmol/L) followed by an insulin rate increase (39,342 pairs; median latency 14 minutes), and hypoglycaemia (<3.9 mmol/L) followed by an insulin rate decrease or stop (2,049 pairs; median latency 8 minutes). Causal inversion was quantified by calculating the proportion of reactive pairs that fall into the same temporal bin, using a maximum response latency of 4 hours. Furthermore, we measured the masking of dysglycaemic events by calculating the proportion of out-of-range events hidden within normoglycaemic bin averages. Finally, we assessed decision distortion by measuring the collapse of two or more insulin adjustments into a single bin.

\subsubsection{Datasets for offline reinforcement learning}

To generate datasets for offline learning, we first trained a proximal policy optimisation (PPO) agent~\cite{schulman2017proximal} to solve the irregular version of each environment, with regular model checkpoints during training. In UVA/Padova, the agent was trained on the training cohort of patients, with each checkpoint evaluated over a fixed number of episodes using the validation cohort of patients. We selected an intermediate checkpoint achieving approximately 70--80\% of the agent's best overall performance to serve as the `expert policy'. This ensures the dataset demonstrates competent but imperfect behaviour, providing room for offline agents to improve.

Using this expert policy, we generated a ground-truth reference dataset by interacting repeatedly with the environment's irregular version (100,000 transitions for LavaGap; 10 million for UVA/Padova). From this reference dataset, we constructed three dataset versions:~\label{reforms3e}\label{reforms3f}

\begin{enumerate}

    \item \textbf{Unprocessed dataset:} Records all state transitions and a mask identifying independent agent decisions, preserving the semi-MDP structure.
    \item \textbf{Interpolated dataset:} Identical to the unprocessed dataset, but treats all transitions as independent decisions. For example, in UVA/Padova, a decision held for 60 minutes is treated as six independent 10-minute decisions.
    \item \textbf{Binned dataset:} Produced via temporal binning. In LavaGap, every second time step is subsampled. In UVA/Padova, time windows (e.g., 2 hours) are used to average observations/actions and sum rewards. Observations are aggregated over the preceding window.\label{reforms6c}

\end{enumerate}\label{reforms4c}

All UVA/Padova datasets had their rewards standardised to zero mean and unit variance using values from the training datasets. No other preprocessing was performed, as the environments already internally normalise their observations.\label{reforms6a}

We perform the same analyses as described for MIMIC-IV on this simulated dataset. The exception to this is the analysis of causal inversion events; unlike MIMIC-IV (where glucose is sampled irregularly), our simulated dataset has regular measurements of blood glucose at every 10-minute interval. We provide a proxy for reactive glucose-insulin pairs by identifying all initial onsets of hypo-/hyperglycaemia (the first glucose reading to cross the target threshold i.e., the earliest opportunity to detect this event), and pair these with the subsequent insulin decision (an increased or decreased dose). Unlike for MIMIC-IV, we use a 2-hour latency period here, as the maximum possible interval between consecutive decisions in our simulated environment. We then report the proportion of these paired events that become causally inverted as a result of temporal binning.

\subsection{Offline training}
\subsubsection{Algorithms}\label{reforms5a}\label{reforms5aiii}\label{reforms5b}

We evaluated the effects of resampling on ORL model performance using three algorithms: behavioural cloning~\cite{kumar2022should} (BC), implicit Q-learning~\cite{kostrikov2022offline} (IQL), and conservative Q-learning~\cite{kumar2020conservative} (CQL)\@. BC serves as a supervised learning baseline. IQL and CQL are state-of-the-art model-free methods that constrain the learned policy to stay within the support of the training data. Detailed implementation details for each algorithm can be found in their respective publications.

Policy networks were trained using negative log-likelihood (for continuous doses in UVA/Padova) or cross-entropy (for discrete actions in LavaGap) losses. All critics using Bellman backups were trained using mean squared error loss, defined using an SMDP framework:

\begin{equation}
    L_Q (\theta) = \frac{1}{2} \mathbb{E}_{(s,a,r,s',\Delta t) \sim \mathcal{D}} [{( r + \gamma^{\Delta t} V_\psi (s') - Q_\theta (s, a))}^2]
\end{equation}

We applied the SMDP framework to each Bellman backup by 1) determining the holding period $\Delta t$ between consecutive decisions; 2) calculating the discounted sum of rewards, $r$, accumulated during this period as defined by Eq.~\ref{eq:smdp-reward-accrued}; and 3) dynamically adjusting the discount factor $\gamma^{\Delta t}$ accordingly. In the case of training on binned and interpolated datasets, $\Delta t$ reduces to a single time step between consecutive states (i.e., $\gamma^{\Delta t} = \gamma$), and the reward $r$ simply becomes the reward for that transition.

\subsubsection{Model Architecture}\label{reforms5aii}

For each environment, all experiments used the same model architecture consisting of a feature encoder and one or more decoding heads, with Leaky ReLU activation for hidden layers.
\begin{itemize}
    \item \textbf{LavaGap:} A convolutional neural network (CNN) feature encoder followed by dense multi-layer perceptron (MLP) decoding heads.
    \item \textbf{UVA/Padova:} An MLP followed by a long short-term memory (LSTM) layer for the feature encoder, and MLPs for each decoding head.
\end{itemize}

In the discrete-action LavaGap environment, all policy and critic networks produced output vectors with dimensionality matching the action space. In the continuous-action UVA/Padova environment, the policy network output the $\alpha$ and $\beta$ parameters of a Beta distribution~\cite{chou2017improving}, scaled to the insulin infusion range (0--0.5 U/min). Critic networks incorporated the action by concatenating it with the latent state representation. During dataset generation, actions were sampled from the PPO action distribution. All evaluation tasks used deterministic outputs corresponding to the action with the highest probability for LavaGap and the distributional mean for UVA/Padova.

\subsection*{Evaluation}\label{reforms3d}\label{reforms5c}\label{reforms5d}\label{reforms7a}\label{reforms7b}\label{reforms7c}
Agarwal et al.~\cite{agarwal2021deep} describe a statistically robust protocol for analysing reinforcement learning experiments, which we follow here. Each experiment (defined by a specific ORL algorithm-dataset pair) consisted of 50 independent training runs with different random seeds.

\begin{itemize}
    \item \textbf{Training:} Offline agents for LavaGap were trained for 10,000 update steps. Agents for UVA/Padova were trained for up to 100,000 update steps, with early stopping based on performance on the validation cohort.
    \item \textbf{Evaluation:} Upon completion of training, the offline agent was deployed in both the regular and irregular versions of the environment (using the test cohort for UVA/Padova). Performance was evaluated over 100 episodes in LavaGap and 30 episodes per test patient in UVA/Padova, with results averaged to obtain a mean score for each run.
    \item \textbf{Statistical Analysis:} Scores from the 50 runs were aggregated using the interquartile mean (IQM) to improve robustness to outlier performances. We computed 95\% confidence intervals using stratified bootstrapping with 100,000 replicates.
\end{itemize}

In addition, for UVA/Padova, we performed off-policy evaluation using fitted Q-evaluation (FQE)~\cite{le2019batch}. FQE is an off-policy evaluation method that estimates the value of a target policy $\pi_e$ by iteratively training a Q-function to minimise the mean squared Bellman error over a fixed dataset $\mathcal{D}$:
\begin{equation}
    L_Q (\theta) = \frac{1}{2} \mathbb{E}_{(s, a, r, s',\Delta t) \sim \mathcal{D}}\left[{\left(r + \gamma^{\Delta t} \mathbb{E}_{a' \sim \pi_e (\cdot|s')} \left[Q_{\hat{\theta}} (s', a')\right] - Q_\theta (s, a)\right)}^2\right]
\end{equation}
For each run, an FQE model was trained for 30,000 update steps to estimate the value of the target policy using the same training cohort dataset employed for policy training. An equivalent dataset of 3 million transitions was generated using the independent (unseen) cohort of test patients, and inference was performed on the initial states of this dataset to produce an FQE score for that run. During FQE training, we used a discount factor $\gamma=1.0$, such that the FQE model output for the initial states approximates the episodic return of the evaluated agent (for a finite horizon $H$), as defined earlier in Eq.~(\ref{eq:semi-mdp-objective}):

\begin{equation}
    \mathbb{E}_{s \sim \mathcal{D};~a' \sim \pi_e (\cdot | s)}[Q_\theta(s_0, a')] \approx \mathbb{E}_{\tau \sim (\pi_e, T, \mu_0)} \left[ \sum_{k=0}^H R(s_k, a_k, s_{k+1}, \Delta t_k) \right]
\end{equation}

This allows for a direct comparison between the predictions from the FQE model and the true performance generated by the model during online evaluation in the simulator. The same statistical analysis was applied to both off-policy and prospective evaluation, including aggregation via the IQM and stratified bootstrapping. The original PPO agent used to generate the dataset served a baseline for each evaluation method, both prospectively (by evaluating PPO in the live environment) and retrospectively (by conducting FQE on the offline dataset behaviour).\label{reforms5f}

\subsubsection{Computing details}\label{reforms2c}

All code and datasets are publicly available at \url{https://github.com/tdgfrost/temporal-resampling-experiments}. ~\label{reforms2b}\label{reforms2d}\label{reforms2e}

Models were implemented in Python (v3.10) using PyTorch~\cite{paszke2019pytorch} (v2.9.0). Custom environments used Gymnasium~\cite{towers2024gymnasium} (0.29.1) and our own Numba-accelerated version of SimGlucose~\cite{xie2018simglucose} (0.2.11). We used the Adam optimiser for all training~\cite{kingma2014adam}, with default PyTorch hyperparameters aside from learning rate.

Critics in IQL and CQL used clipped double Q-networks. Target networks were updated using Polyak averaging with $\alpha = 0.005$. A full table of hyperparameters is provided in Table~\ref{tab:hyperparameters}. For the IQL and CQL hyperparameters, a preliminary hyperparameter sweep was performed to identify stable optima consistent across all three temporal resampling regimes, ensuring comparisons were not biased by dataset-specific tuning.\label{reforms5ei}\label{reforms5eii} A REFORMS checklist~\cite{kapoor2024reforms} is provided in Supplementary Table~\ref{tab:reforms}.

\begin{table}[htbp]
\centering
\caption{Hyperparameters for LavaGap and UVA/Padova environments.}\label{tab:hyperparameters}
\small 
\begin{tabular}{l l l l}
\toprule
\textbf{Hyperparameters} & \textbf{LavaGap} & \textbf{UVA/Padova} & \textbf{UVA/Padova (FQE)} \\ \midrule
Learning rate & $1 \cdot 10^{-3}$ & $3 \cdot 10^{-4}$ & $3 \cdot 10^{-4}$ \\
Batch size & 64 & 1024 & 256 \\
Hidden dimensions & 64 & 128 & 64 \\
$\gamma$ --- discount factor & 0.99 & 0.99 & 1.0 \\
$\tau$ --- expectile (IQL) & 0.8 & 0.9 & --- \\
$\beta$ --- temperature (IQL) & 10 & 10 & --- \\
$\alpha$ --- reg.\ term (CQL) & 1.0 & 1.0 & --- \\
$\alpha$ --- entropy term (CQL) & 0.2 & 0.0 & --- \\ \bottomrule
\end{tabular}\label{reforms5eiii}
\end{table}

\section*{Author Contribution}
All authors have read and approved this manuscript. TF and HV designed the experiments. TF wrote the software code, which HV checked. TF analysed the experimental results. TF and HV wrote the manuscript. SH provided supervision.

\section*{Acknowledgements}
TF was funded by the Engineering and Physical Sciences Research Council as part of the UK Research and Innovation
Centre for Doctoral Training in AI for Healthcare (grant EP/S021612/1). HV was funded by the National Institute for Health and Care Research through an Academic Clinical Fellowship (ACF-2023-18-012). SH supported the research as a researcher based at the National Institute for Health and Care Research (NIHR) University College London Hospitals (UCLH) Biomedical Research Centre (BRC)\@. Funders played no role in study design, data collection and analysis, decision to publish, or the writing of this manuscript. The views expressed in the text are those of the authors and not necessarily those of the funders. We also gratefully acknowledge the facilities provided by University College London (UCL), which helped enable this research. Lastly, we thank Ahmed Al-Hindawi, Simon Ellershaw, Jennifer Hunter, Aasiyah Rashan, Dan Stein, Dylan Whitaker, and Matt Wilson for their valued input during the preparation of this manuscript.

\section*{Ethics Declarations}
All authors declare no competing interests. No ethical approval was required for this research.

\nocite{nemati2016optimal}
\nocite{lin2018deep}
\nocite{wang2018supervised}
\nocite{liu2019learning}
\nocite{lopez2019deep}
\nocite{yu2019inverse}
\nocite{futoma2020identifying}
\nocite{yu2020supervised}
\nocite{peine2021development}
\nocite{eghbali2021patient}
\nocite{sun2021personalized}
\nocite{yala2022optimizing}
\nocite{guo2022learning}
\nocite{li2022electronic}
\nocite{nanayakkara2022unifying}
\nocite{prasad2022guiding}
\nocite{su2022establishment}
\nocite{huang2022reinforcement}
\nocite{denhengst2024guideline}
\nocite{lee2024reinforcement}
\nocite{zhang2024optimizing}
\nocite{desman2025distributional}
\nocite{kalimouttou2025optimal}
\nocite{lee2025optimizing}
\bibliography{references}

\newpage

\appendix
\pagestyle{fancy}
\section{Supplementary Materials}
\setcounter{figure}{0}
\setcounter{table}{0}

\renewcommand{\thefigure}{S\arabic{figure}}
\renewcommand{\thetable}{S\arabic{table}}
\renewcommand{\theHfigure}{S\arabic{figure}}
\renewcommand{\theHtable}{S\arabic{table}}

\flushbottom
\vspace{\stretch{1}}

\begin{figure}[htbp]
    \centering
    \includegraphics[width=0.45\textwidth]{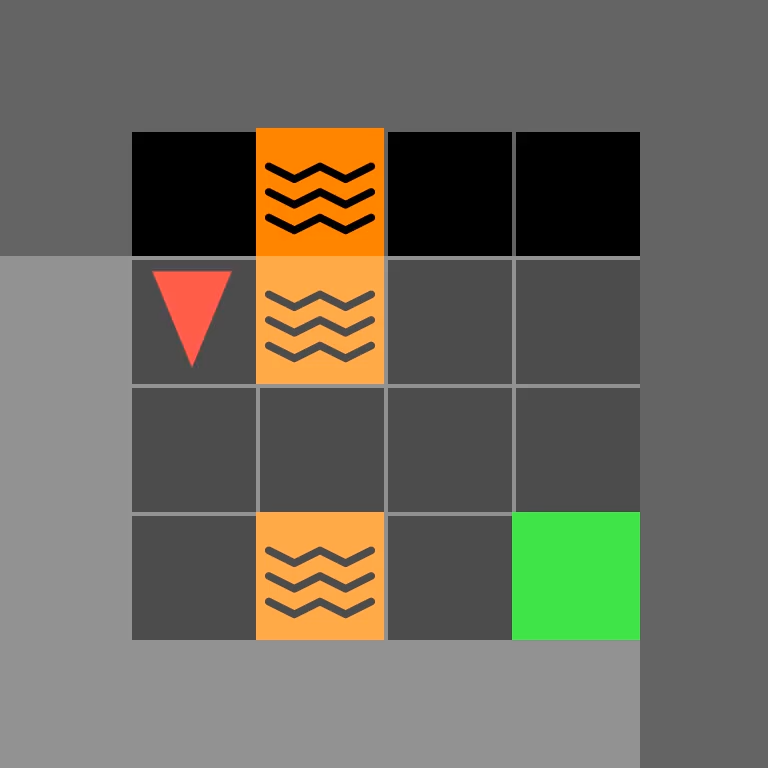}
    \caption{\textbf{LavaGap Environment.} To reach the green goal square, the agent (red triangle) must navigate the grid without contacting the lava (orange squares). The model uses partial observations (light grey) to calculate the optimal path. The episode ends with a reward of 1 if the agent reaches the goal, or 0 if it hits lava or exceeds the maximum number of steps.}\label{fig:lavagap_example}
\end{figure}

\vspace{\stretch{1}}

\begin{figure}[htbp]
    \centering
    \includegraphics[width=\textwidth]{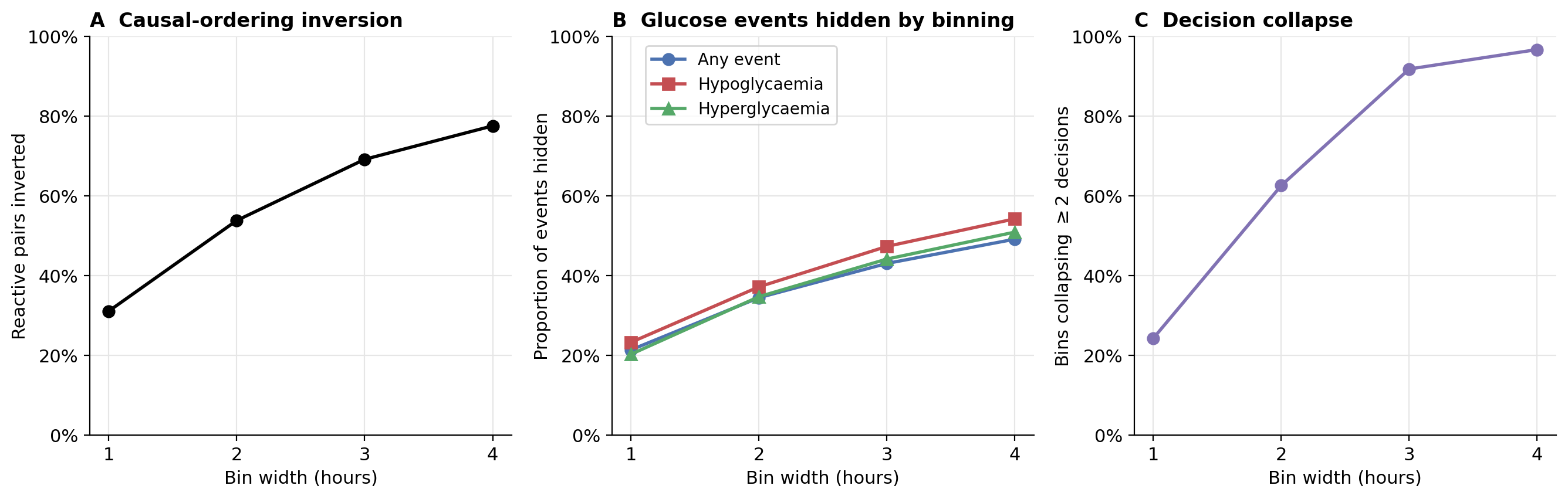}
    \caption{\textbf{Quantification of artefacts in UVA/Padova following temporal binning.} A) The proportion of causally inverted artefacts from insulin adjustments paired with the preceding onset of hypo-/hyperglycaemia, as a result of both events sharing the same time window. B) The proportion of hypo- and hyperglycaemic events that are masked as a result of averaging blood glucose levels within each bin window. C) The proportion of bins containing 2 or more insulin adjustment decisions (which are then aggregated into a single action).}\label{fig:simglucose_binning_metrics}
\end{figure}

\raggedbottom

\begin{figure}[htbp]
    \centering
    \includegraphics[width=\textwidth]{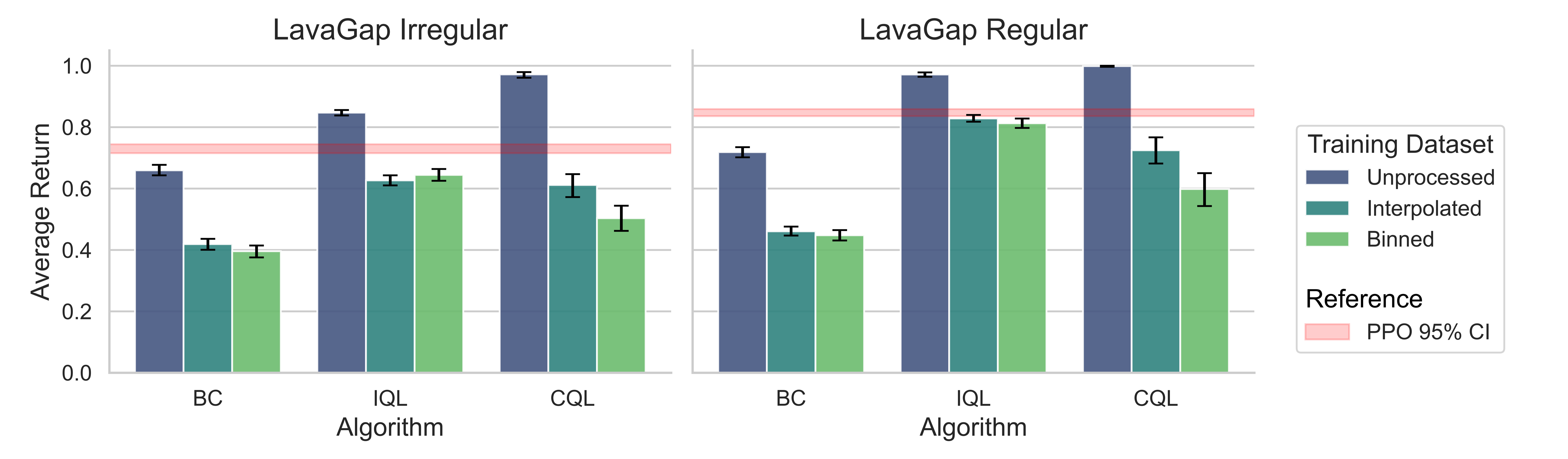}
    \caption{\textbf{Results from LavaGap.} Offline reinforcement learning agents (BC, IQL, CQL) demonstrate universal reductions in performance when trained on temporally resampled datasets (interpolated, binned) compared to unprocessed datasets with natural timings. In combination with Fig~\ref{fig:padova_results}, this effect appears to generalise across both discrete and continuous action domains. Error bars and shaded regions represent 95\% confidence intervals.}\label{fig:lavagap_results}
\end{figure}

\begin{figure}[htbp]
    \centering
    \includegraphics[width=\textwidth]{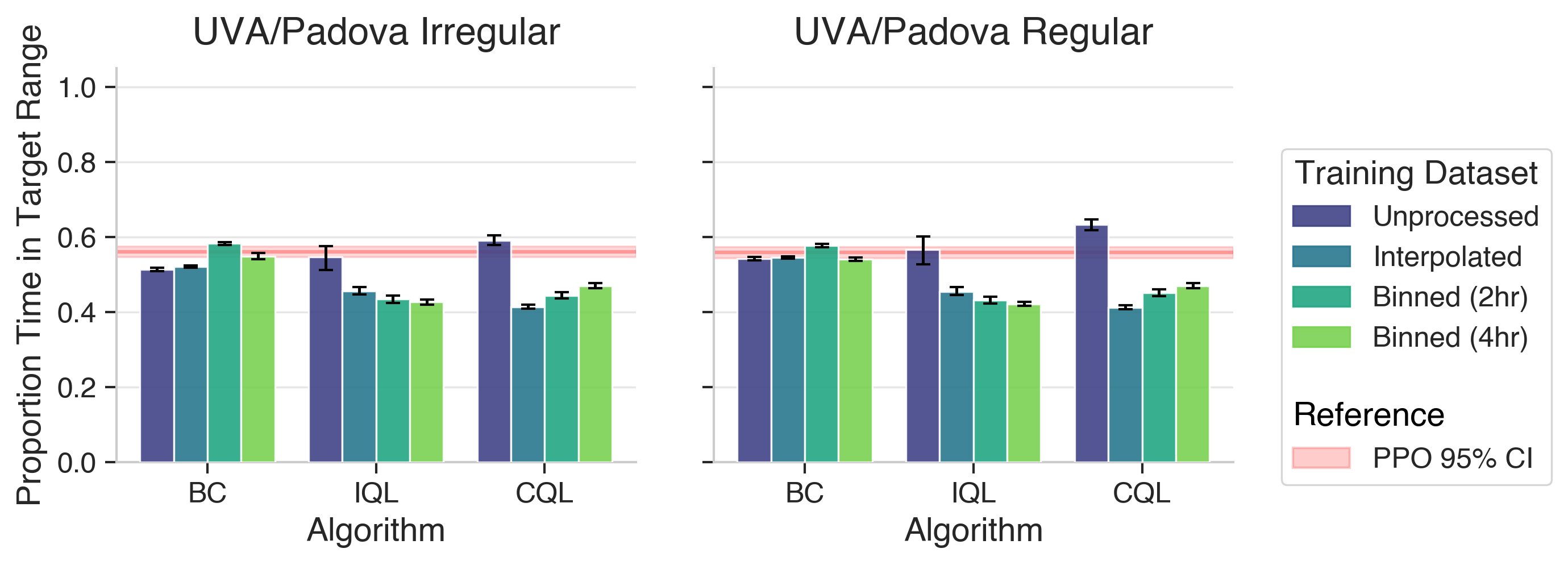}
    \caption{\textbf{Time in target range performance in the UVA/Padova simulator.} Proportion of time spent within the target glucose range for each offline reinforcement learning agent (BC, IQL, CQL). Under both irregular (left) and regular (right) sampling, IQL and CQL achieve peak performance using unprocessed datasets, matching or exceeding the online PPO baseline. However, training on temporally resampled datasets (interpolated and binned) leads to a consistent degradation in performance for both algorithms. Error bars and the shaded horizontal region represent 95\% confidence intervals.}\label{fig:time_in_target_range}
\end{figure}

\begin{figure}[htbp]
    \centering
    \includegraphics[width=\textwidth]{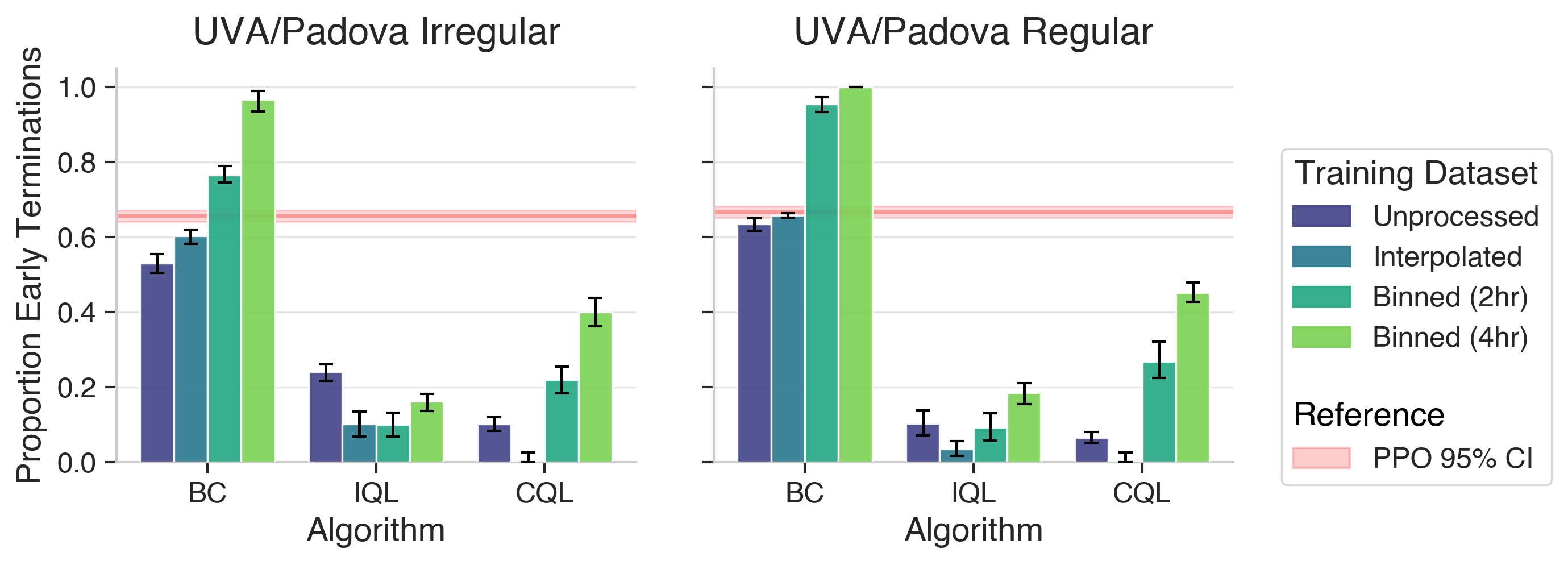}
    \caption{\textbf{Early termination rates in the UVA/Padova simulator}: Proportion of episodes ending in early termination for offline reinforcement learning agents (BC, IQL, CQL) under irregular (left) and regular (right) sampling conditions. BC shows a high propensity for early termination, approaching 1.0 when trained on binned datasets for both environments. IQL and CQL both lead to a lower proportion of episodes with early termination, with a trend towards the lowest values from interpolated data and the highest values from binned data. Error bars and shaded regions represent 95\% confidence intervals, with the pink band indicating the PPO reference baseline.}\label{fig:proportion_early_termination}
\end{figure}

\begin{figure}[htbp]
    \centering
    \includegraphics[width=0.8\textwidth]{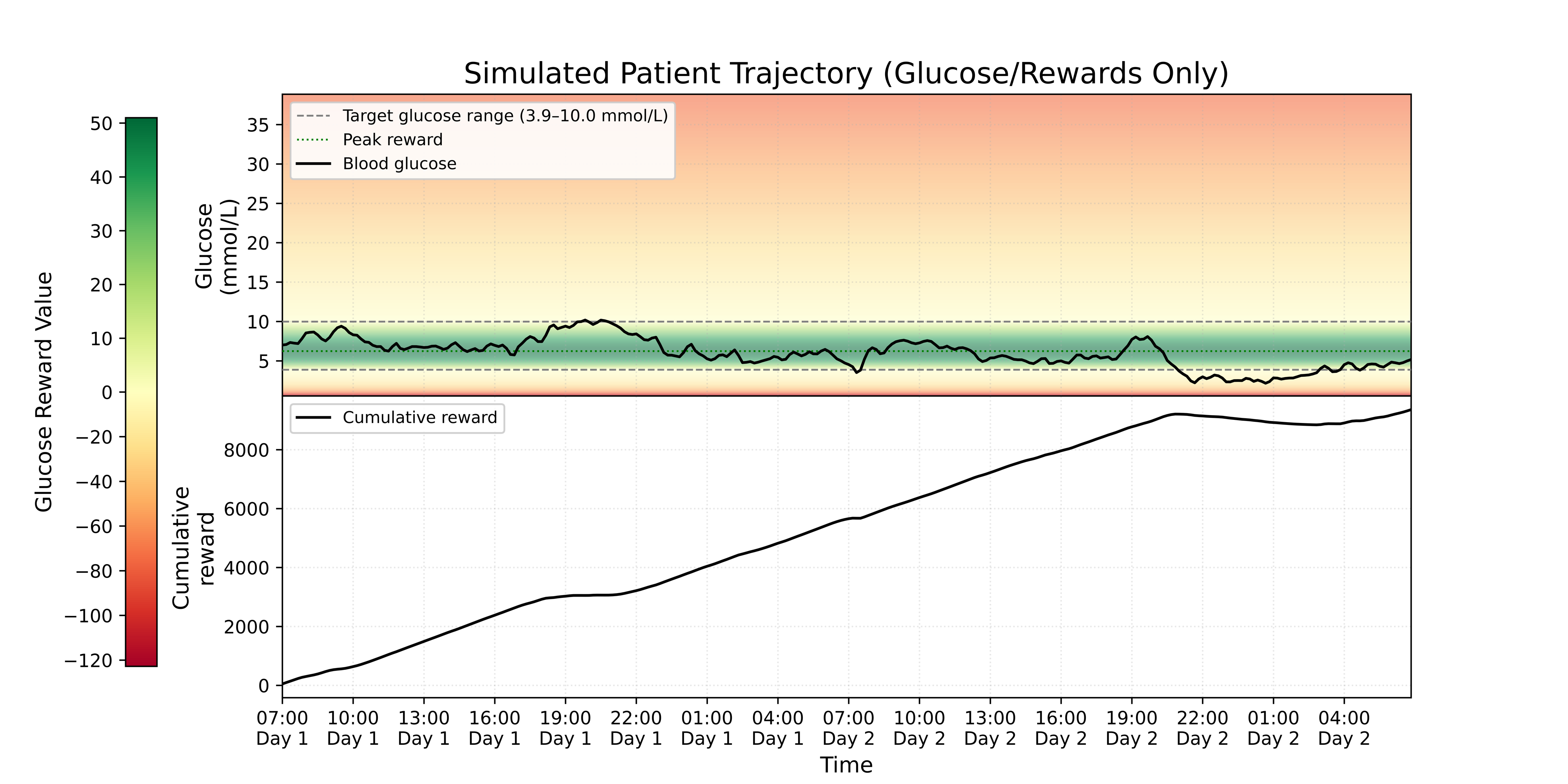}
    \caption{\textbf{Example patient trajectory in the UVA/Padova simulator (glucose and rewards only).} The figure shows how the per-step and cumulative rewards change for different values of glucose in the UVA/Padova T1DM simulator over the course of an example patient trajectory.}\label{fig:simglucose_example_extended}
\end{figure}

\renewcommand{\arraystretch}{1.2}

\begin{landscape}
\newpage
\nexttablelabel{tab:lit-review}

\footnotesize
\newcounter{rowcnt}
\newcommand{\rn}{\stepcounter{rowcnt}\textbf{\arabic{rowcnt}}}
\begin{longtable}{c P{1.4cm} P{2.4cm} P{5.8cm} P{2cm} P{2.5cm} P{2.5cm} P{1.6cm}}
    \caption{\textbf{Literature Review}: Summary of relevant literature from 2016 to 2026 for offline reinforcement learning using electronic health record data, including choice of resampling interval where applicable.}\\
    \toprule
    \textbf{Paper} & \textbf{Date} & \textbf{Authors} & \textbf{Title} & \textbf{Data} & \textbf{Model} & \textbf{Evaluation} & \textbf{Intervals} \\
    \midrule
\endfirsthead
    \toprule
    \textbf{Paper} & \textbf{Date} & \textbf{Authors} & \textbf{Title} & \textbf{Data} & \textbf{Model} & \textbf{Evaluation} & \textbf{Intervals} \\
    \midrule
\endhead

\textbf{\rn} & 2016 Aug & Nemati et al.~\cite{nemati2016optimal} & Optimal medication dosing from suboptimal clinical examples & MIMIC-II & Fitted Q-Iteration (model-free) & Retrospective evaluation only & 1 hour \\
\textbf{\rn} & 2017 Aug & Prasad et al.~\cite{prasad2017reinforcement} & A reinforcement learning approach to weaning of mechanical ventilation in intensive care units & MIMIC-III & Fitted Q-Iteration (model-free) & Retrospective evaluation only & 10 minutes \\
\textbf{\rn} & 2017 Nov & Raghu et al.~\cite{raghu2017continuous} & Continuous state-space models for optimal sepsis treatment: a deep reinforcement learning approach & MIMIC-III & Dueling Double Deep Q-Network (model-free) & Retrospective evaluation only & 4 hours \\
\textbf{\rn} & 2018 Jul & Lin et al.~\cite{lin2018deep} & A deep deterministic policy gradient approach to medication dosing and surveillance in the ICU & MIMIC-III, Emory University & Deep Deterministic Policy Gradient (model-free) & Retrospective evaluation only & Unspecified (3 hours?) \\
\textbf{\rn} & 2018 Jul & Wang et al.~\cite{wang2018supervised} & Supervised reinforcement learning with recurrent neural network for dynamic treatment recommendation & MIMIC-III & Supervised RL with Recurrent Neural Network (model-free) & Retrospective evaluation only & 24 hours \\
\textbf{\rn} & 2018 Nov & Komorowski et al.~\cite{komorowski2018artificial} & The artificial intelligence clinician learns optimal treatment strategies for sepsis in intensive care & MIMIC-III, eICU & Policy Iteration (model-based) & Retrospective evaluation only & 4 hours \\
\textbf{\rn} & 2018 Dec & Peng et al.~\cite{peng2018improving} & Improving sepsis treatment strategies by combining deep and kernel-based reinforcement learning & MIMIC-III & Dueling Double Deep Q-Network (model-free) & Retrospective evaluation only & 4 hours \\
\textbf{\rn} & 2019 Feb & Liu et al.~\cite{liu2019learning} & Learning the dynamic treatment regimes from medical registry data through deep Q-network & CIBMTR & Deep Q-Network (model-free) & Retrospective evaluation only & 5 total bins \\
\textbf{\rn} & 2019 Jul & Lopez et al.~\cite{lopez2019deep} & Deep reinforcement learning for optimal critical care pain management with morphine using dueling double-deep Q networks & MIMIC-III & Dueling Double Deep Q-Network (model-free) & Retrospective evaluation only & 1 hour \\
\textbf{\rn} & 2019 Apr & Yu et al.~\cite{yu2019inverse} & Inverse reinforcement learning for intelligent mechanical ventilation and sedative dosing in intensive care units & MIMIC-III & Inverse RL with Fitted Q-Iteration (model-free) & Retrospective evaluation only & 10 minutes \\
\textbf{\rn} & 2020 May & Futoma et al.~\cite{futoma2020identifying} & Identifying distinct, effective treatments for acute hypotension with SODA-RL\@: Safely optimized diverse accurate reinforcement learning & MIMIC-III & SODA-RL (model-free) & Retrospective evaluation only & 1 hour \\
\textbf{\rn} & 2020 July & Tang et al.~\cite{tang2020clinician} & Clinician-in-the-loop decision making: reinforcement learning with near-optimal set-valued policies & MIMIC-III & Set-Valued Policy (model-free) & Retrospective evaluation only & 4 hours \\
\textbf{\rn} & 2020 Jul & Yu et al.~\cite{yu2020supervised} & Supervised-actor-critic reinforcement learning for intelligent mechanical ventilation and sedative dosing in intensive care units & MIMIC-III & Supervised Actor-Critic (model-free) & Retrospective evaluation only & 10 minutes \\
\textbf{\rn} & 2020 Nov & Lu et al.~\cite{lu2020deep} & Is deep reinforcement learning ready for practical applications in healthcare? A sensitivity analysis of duel-DDQN for hemodynamic management in sepsis patients & MIMIC-III & Dueling Double Deep Q-Network (model-free) & Retrospective evaluation only & 1 hour \\
\textbf{\rn} & 2020 Dec & Killian et al.~\cite{killian2020empirical} & An empirical study of representation learning for reinforcement learning in healthcare & MIMIC-III & Batch-Constrained Q-Learning (model-free) & Retrospective evaluation only & 4 hours \\
\textbf{\rn} & 2021 Feb & Peine et al.~\cite{peine2021development} & Development and validation of a reinforcement learning algorithm to dynamically optimize mechanical ventilation in critical care & MIMIC-III, eICU & Q-Learning (model-free) & Retrospective evaluation only & 4 hours \\
\textbf{\rn} & 2021 Feb & Roggeveen et al.~\cite{roggeveen2021transatlantic} & Transatlantic transferability of a new reinforcement learning model for optimizing haemodynamic treatment for critically ill patients with sepsis & MIMIC-III, Amsterdam UMCdb & Dueling Double Deep Q-Network (model-free) & Retrospective evaluation only & 4 hours \\
\textbf{\rn} & 2021 Mar & Eghbali et al.~\cite{eghbali2021patient} & Patient-specific sedation management via deep reinforcement learning & MIMIC-IV & Deep Deterministic Policy Gradient (model-free) & Retrospective evaluation only & 1 hour \\
\textbf{\rn} & 2021 Aug & Sun et al.~\cite{sun2021personalized} & Personalized vital signs control based on continuous action-space reinforcement learning with supervised experience & MIMIC-III & Deep Deterministic Policy Gradient (model-free) & Retrospective evaluation only & 1 hour \\
\textbf{\rn} & 2021 Nov & Liu et al.~\cite{liu2021combining} & Combining model-based and model-free reinforcement learning policies for more efficient sepsis treatment & MIMIC-III & Policy Iteration + Dueling Deep Q-Network (model-based and model-free) & Retrospective evaluation only & 4 hours \\
\textbf{\rn} & 2021 Dec & Fatemi et al.~\cite{fatemi2021medical} & Medical dead-ends and learning to identify high-risk states and treatments & MIMIC-III & Customised Double Deep Q-Networks (model-free) & Retrospective evaluation only & 4 hours \\
\textbf{\rn} & 2021 Dec & Satija et al.~\cite{satija2021multi} & Multi-objective SPIBB\@: Seldonian offline policy improvement with safety constraints in finite MDPs & MIMIC-III & Safe Policy Improvement with Baseline Bootstrapping (model-based) & Retrospective evaluation only & 4 hours \\
\textbf{\rn} & 2022 Jan & Yala et al.~\cite{yala2022optimizing} & Optimizing risk-based breast cancer screening policies with reinforcement learning & Massachusetts General Hospital & Envelope Q-Learning (model-free and model-based) & Retrospective evaluation only & 6 months \\
\textbf{\rn} & 2022 Feb & Guo et al.~\cite{guo2022learning} & Learning dynamic treatment strategies for coronary heart diseases by artificial intelligence: real-world data-driven study & MIMIC-III & Deep Deterministic Policy Gradient (model-free) & Retrospective evaluation only & 24 hours \\
\textbf{\rn} & 2022 Feb & Li et al.~\cite{li2022electronic} & Electronic health records based reinforcement learning for treatment optimizing & MIMIC-III & Deep Q-Network + Model-Based RL (model-free and model-based) & Retrospective evaluation only & 3 hours \\
\textbf{\rn} & 2022 Feb & Nanayakkara et al.~\cite{nanayakkara2022unifying} & Unifying cardiovascular modelling with deep reinforcement learning for uncertainty aware control of sepsis treatment & MIMIC-III & Distributional RL (model-free) & Retrospective evaluation only & 1 hour \\
\textbf{\rn} & 2022 Apr & Prasad et al.~\cite{prasad2022guiding} & Guiding efficient, effective, and patient-oriented electrolyte replacement in critical care & UPHS, MIMIC-IV & Fitted Q-Iteration (model-free) & Retrospective evaluation only & 6 hours \\
\textbf{\rn} & 2022 Apr & Su et al.~\cite{su2022establishment} & Establishment and implementation of potential fluid therapy balance strategies for ICU sepsis patients based on reinforcement learning & PICMISD & Deep Q-Network (model-free) & Retrospective evaluation only & 6 hours \\
\textbf{\rn} & 2022 Aug & Huang et al.~\cite{huang2022reinforcement} & Reinforcement learning for sepsis treatment: a continuous action space solution & MIMIC-III & Deep Deterministic Policy Gradient (model-free) & Retrospective evaluation only & 4 hours \\
\textbf{\rn} & 2022 Sep & Shiranthika et al.~\cite{shiranthika2022supervised} & Supervised optimal chemotherapy regimen based on offline reinforcement learning & Local data & Conservative Q-Learning (model-free) & Retrospective evaluation only & \textbf{Already regular} \\
\textbf{\rn} & 2023 Feb & Wu et al.~\cite{wu2023value} & A value-based deep reinforcement learning model with human expertise in optimal treatment of sepsis & MIMIC-III & Weighted Dueling Double Deep Q-Network (model-free) & Retrospective evaluation only & 1, 2, 4, 6, and 8 hours \\
\textbf{\rn} & 2023 May & Gao et al.~\cite{gao2023offline} & Offline learning of closed-loop deep brain stimulation controllers for Parkinson disease treatment & Local & Deep Deterministic Policy Gradient (model-free) & \textbf{Retrospective evaluation + clinical trial} & \textbf{Already regular} \\
\textbf{\rn} & 2023 May & Liang et al.~\cite{liang2023treatment} & The treatment of sepsis: an episodic memory-assisted deep reinforcement learning approach & MIMIC-III & Dueling Double Deep Q-Network (model-free) & Retrospective evaluation only & 4 hours \\
\textbf{\rn} & 2023 June & Kondrup et al.~\cite{kondrup2023towards} & Towards safe mechanical ventilation treatment using deep offline reinforcement learning & MIMIC-III & Conservative Q-Learning (model-free) & Retrospective evaluation only & 4 hours \\
\textbf{\rn} & 2023 Oct & Wang et al.~\cite{wang2023optimized} & Optimized glycemic control of type 2 diabetes with reinforcement learning: a proof-of-concept trial & Zhongshan Hospital, Qingpu Hospital & Customised Model-based RL (model-based) & \textbf{Retrospective evaluation + clinical trial} & 7 daily bins \\
\textbf{\rn} & 2023 Oct & Zhu et al.~\cite{zhu2023offline} & Offline deep reinforcement learning and off-policy evaluation for personalized basal insulin control in type 1 diabetes & UVA/Padova T1DM, OhioT1DM & TD3-BC (model-free) & Retrospective evaluation + simulated trial & \textbf{Already regular} \\
\textbf{\rn} & 2024 Jan & den Hengst et al.~\cite{denhengst2024guideline} & Guideline-informed reinforcement learning for mechanical ventilation in critical care & MIMIC-III & Q-Learning, Imitation Learning (model-free) & Retrospective evaluation only & 4 hours \\
\textbf{\rn} & 2024 Jan & Lee et al.~\cite{lee2024reinforcement} & Reinforcement learning model for optimizing dexmedetomidine dosing to prevent delirium in critically ill patients & SNUH & Conservative Q-Learning (model-free) & Retrospective evaluation only & 6 hours \\
\textbf{\rn} & 2024 Feb & Petch et al.~\cite{petch2024optimizing} & Optimizing warfarin dosing for patients with atrial fibrillation using machine learning & ARISTOTLE, ENGAGE AF-TIMI, ROCKET AF, RE-LY & Batch-Constrained Q-Learning (model-free) & Retrospective evaluation only & \textbf{Not used} \\
\textbf{\rn} & 2024 Mar & Zhang et al.~\cite{zhang2024optimizing} & Optimizing sepsis treatment strategies via a reinforcement learning model & MIMIC-III & Dueling Double Deep Q-Network (model-free) & Retrospective evaluation only & 2 hours \\
\textbf{\rn} & 2024 Apr & Job et al.~\cite{job2024optimal} & Optimal treatment strategies for critical patients with deep reinforcement learning & MIMIC-III & Dueling Double Deep Q-Network (model-free) & Retrospective evaluation only & \textbf{Not used} \\
\textbf{\rn} & 2024 Aug & Beolet et al.~\cite{beolet2024end} & End-to-end offline reinforcement learning for glycemia control & UVA/Padova T1DM & BCQ, TD3-BC, CQL (model-free) & Simulated trial only & 1 hour \\
\textbf{\rn} & 2025 Feb & Ghasemi et al.~\cite{ghasemi2025personalized} & Personalized decision making for coronary artery disease treatment using offline reinforcement learning & APPROACH & Q-Learning, Deep Q-Learning, Conservative Q-Learning (model-free) & Retrospective evaluation only & \textbf{Not used} \\
\textbf{\rn} & 2025 Feb & Tu et al.~\cite{tu2025offline} & Offline safe reinforcement learning for sepsis treatment: tackling variable-length episodes with sparse rewards & MIMIC-III & Conservative Q-Learning, Double Deep Q-Network (model-free) & Retrospective evaluation only & 4 hours \\
\textbf{\rn} & 2025 May & Desman et al.~\cite{desman2025distributional} & A distributional reinforcement learning model for optimal glucose control after cardiac surgery & MIMIC-IV, eICU & Distributional RL + Conservative Q-Learning (model-free) & Retrospective evaluation only & 1 hour \\
\textbf{\rn} & 2025 May & Kalimouttou et al.~\cite{kalimouttou2025optimal} & Optimal vasopressin initiation in septic shock: the OVISS reinforcement learning study & MIMIC-IV, eICU, UPMC, UCSF & Fitted Q-Iteration (model-free) & Retrospective evaluation only & 1 hour \\
\textbf{\rn} & 2025 Aug & Zhou et al.~\cite{zhou2025optimizing} & Optimizing long term disease prevention with reinforcement learning: a framework for precision lipid control & HKHA & Policy Iteration (model-based) & Retrospective evaluation only & \textbf{Not used} \\
\textbf{\rn} & 2025 Sep & Fang et al.~\cite{fang2025offline} & Offline inverse constrained reinforcement learning for safe-critical decision making in healthcare & MIMIC-III & Constraint Transformer (model-based) & Retrospective evaluation only & 4 hours \\
\textbf{\rn} & 2025 Oct & Lee et al.~\cite{lee2025optimizing} & Optimizing Loop Diuretic Treatment for Mortality Reduction in Patients With Acute Dyspnea Using a Practical Offline Reinforcement Learning Pipeline for Health Care: Retrospective Single-Center Simulation Study & Michigan Medicine & Batch-Constrained Q-Learning, Pessimistic MDP (model-free and model-based) & Retrospective evaluation only & 24 hours \\
\textbf{\rn} & 2026 Feb & Fan et al.~\cite{fan2026reinforcement} & Reinforcement learning-based digital therapeutic intervention for postprostatectomy incontinence: development and pilot feasibility study & Local & Q-learning (model-free) & \textbf{Retrospective evaluation + clinical trial} & 1 hour \\

\bottomrule
\end{longtable}

\end{landscape}

\newpage
\nexttablelabel{tab:reforms}

\small
\setcounter{table}{1}
\begin{longtable}{c P{6.5cm} c P{6.5cm}}
\caption{\textbf{REFORMS checklist}: As per Kapoor et al.~\cite{kapoor2024reforms}.}\\
\toprule
\rowcolor[rgb]{0.753,0.753,0.753}
\textbf{Item} & \textbf{Description} & \textbf{Page} & \textbf{Notes} \\
\midrule
\endfirsthead
\toprule
\rowcolor[rgb]{0.753,0.753,0.753}
\textbf{Item} & \textbf{Description} & \textbf{Page} & \textbf{Notes} \\
\midrule
\endhead
\multicolumn{4}{l}{\textbf{Module 1: Study goals}} \\
1a & Population or distribution about which the scientific claim is made. &~\pageref{reforms1a} & Offline reinforcement learning on any temporally resampled healthcare data. \\
1b & Motivation for choosing this population or distribution (1a.). &~\pageref{reforms1b} & Explore risks to patients from decision models trained on resampled (rather than unprocessed) data. \\
1c & Motivation for the use of ML methods in the study. &~\pageref{reforms1c} & Clinical reinforcement learning could improve healthcare decision models, but often mandates an offline training approach.\\
\midrule

\multicolumn{4}{l}{\textbf{Module 2: Computational reproducibility}} \\
2a & Dataset used for training and evaluating the model along with link or DOI to uniquely identify the dataset. &~\pageref{reforms2a}
& Datasets were generated uniquely by interacting with the LavaGap and UVA/Padova environments. Available at the GitHub repo below. \\
2b & Code used to train and evaluate the model and produce the results reported in the paper along with link or DOI to uniquely identify the version of the code used. &~\pageref{reforms2b}
& Code, datasets, instructions all publicly available at the associated GitHub repo. \\
2c & Description of the computing infrastructure used. &~\pageref{reforms2c} & Python~(v3.10), PyTorch~(v2.9.0), Gymnasium~(0.29.1), Numba-accelerated SimGlucose. (See Methods).\\
2d & README file which contains instructions for generating the results using the provided dataset and code. &~\pageref{reforms2d} 
& Code, datasets, instructions all publicly available at the associated GitHub repo.\\
2e & Reproduction script to produce all results reported in the paper. &~\pageref{reforms2e} 
& Code, datasets, instructions all publicly available at the associated GitHub repo.\\
\midrule

\multicolumn{4}{l}{\textbf{Module 3: Data quality}} \\
3a & Sources of data, separately for the training and evaluation datasets (if applicable), along with the time when the datasets are collected, the source and process of ground-truth annotations, and other data documentation. &~\pageref{reforms3a} & Expert PPO policy used to generate 100k transitions (LavaGap) and 10M~transitions (UVA/Padova), which are processed as described in the Methods.\\
3b & Distribution or set from which the dataset is sampled (i.e., the sampling frame). &~\pageref{reforms3b} & As described in the methods.\\
3c & Justification for why the dataset is useful for the modeling task at hand. &~\pageref{reforms3c} & Generating from a clinically validated simulator enables quick, safe prospective evaluation of the models. \\
3d & The definition of the outcome variable of the model along with descriptive statistics, if applicable. &~\pageref{reforms3d} & The interquartile mean of episode returns with 95\% confidence intervals.\\
3e & Number of samples in the dataset. &~\pageref{reforms3e} & 100,000 transitions (LavaGap); 10,000,000 transitions (UVA/Padova). \\
3f & Percentage of missing data, split by class for a categorical outcome variable. &~\pageref{reforms3f} & 0\% missing data (generated from simulated environments). \\
3g & Justification for why the distribution or set from which the dataset is drawn (3b.) is representative of the one about which the scientific claim is being made (1a.). &~\pageref{reforms3g} & Datasets are drawn from environments with irregular decision intervals. They can then be resampled prior to training. \\
\midrule

\multicolumn{4}{l}{\textbf{Module 4: Data preprocessing}} \\
4a & Identification of whether any samples are excluded with a rationale for why they are excluded. &~\pageref{reforms4a} & UVA/Padova: 12 patients excluded from training --- 6 for validation, 6 for testing.\\
4b & How impossible or corrupt samples are dealt with. &~\pageref{reforms4b} & N/A for LavaGap. For UVA/Padova, episodes terminated early if glucose $<10$~mg/dL or $>600$~mg/dL (with a $-10,000$ penalty). \\
4c & All transformations of the dataset from its raw form (3a.) to the form used in the model, for instance, treatment of missing data and normalization. &~\pageref{reforms4c} & Temporal binning (2h/4h), interpolation (10 mins), and reward standardization to zero mean/unit variance. \\
\midrule

\multicolumn{4}{l}{\textbf{Module 5: Modeling}} \\
5a & Detailed descriptions of all models trained, including: &~\pageref{reforms5a} & BC, IQL, and CQL algorithms trained using identical neural networks across three dataset variants (unprocessed, binned, interpolated).\\
   & All features used in the model (including any feature selection). &~\pageref{reforms5ai} & LavaGap: partial grid pixels. UVA/Padova: blood glucose, insulin rate, time of day, and carbohydrate intake.\\
   & Types of models implemented (e.g., Random Forests, Neural Networks). &~\pageref{reforms5aii}& LavaGap: CNN encoder with MLP heads. UVA/Padova: MLP + LSTM encoder with MLP heads.\\
   & Loss function used. &~\pageref{reforms5aiii}& Mean squared error for value/critic estimation, modified with SMDP for unprocessed datasets. Cross-entropy (LavaGap) or negative log-likelihood (UVA/Padova) for policy training.\\
5b & Justification for the choice of model types implemented. &~\pageref{reforms5b} & IQL/CQL selected as state-of-the-art model-free offline RL\@; BC used as a supervised learning baseline.\\
5c & Method for evaluating the models reported in the paper, including details of train-test splits or cross-validation folds. &~\pageref{reforms5c} & 50 independent runs per experiment. UVA/Padova uses a 6-patient test cohort (30 episodes/patient); LavaGap uses 100 episodes.\\
5d & Method for selecting the models reported in the paper. &~\pageref{reforms5d} & UVA/Padova: early stopping based on validation cohort performance to avoid overfitting. LavaGap: trained for fixed 10,000 update steps.\\
5e & For the models reported in the paper, specify details about the hyperparameter tuning: & & \\
   & Range of hyper-parameters used and a justification for why this range is reasonable. &~\pageref{reforms5ei} & Parameters were explored via a preliminary search to identify regions of stable optima across all three temporal resampling regimes.\\
   & Method to select the best hyper-parameter configuration. &~\pageref{reforms5eii} & The final configuration was selected by confirming consistent stable performance across all dataset variants, ensuring comparisons were not biased by dataset-specific tuning.\\
   & Specification of all hyper-parameters used to generate results reported in the paper. &~\pageref{reforms5eiii}& Described in Table~\ref{tab:hyperparameters} \\
5f & Justification that model comparisons are against appropriate baselines. &~\pageref{reforms5f} & Compared against the PPO expert baseline (which generated the original training datasets).\\
\midrule

\multicolumn{4}{l}{\textbf{Module 6: Data leakage}} \\
6a & Justification that pre-processing (Section 4) and modeling (Section 5) steps only use information from the training dataset (and not the test dataset). &~\pageref{reforms6a} & All training data and normalization parameters were calculated strictly using the training patient cohort (18-patient cohort for UVA/Padova).\\
6b & Methods to address dependencies or duplicates between the training and test datasets (e.g.\ different samples from the same patients are kept in the same dataset partition). &~\pageref{reforms6b} 
&  Data leakage was prevented by partitioning the 30 virtual patients in UVA/Padova into mutually exclusive training (18), validation (6), and test (6) cohorts. 
\\
6c & Justification that each feature or input used in the model is legitimate for the task at hand and does not lead to leakage. &~\pageref{reforms6c} & For UVA/Padova, all observations (blood glucose, carbohydrate, insulin, time of day) are from the past/present only, with no leakage of future information. Any aggregation of observations was done using the preceding window only (to avoid look-ahead bias). LavaGap uses the immediately observable pixels only. \\
\midrule

\multicolumn{4}{l}{\textbf{Module 7: Metrics and uncertainty}} \\
7a & All metrics used to assess and compare model performance (e.g., accuracy, AUROC etc.). Justify that the metric used to select the final model is suitable for the task. &~\pageref{reforms7a} & Primary metric: cumulative undiscounted return (interquartile mean), representing the true live performance of the model in the environment. FQE used for off-policy estimation of the same metric.
\\
7b & Uncertainty estimates (e.g., confidence intervals, standard deviations), and details of how these are calculated. &~\pageref{reforms7b} & 95\%~confidence intervals calculated using the procedure described in the Methods. \\
7c & Justification for the choice of statistical tests (if used) and a check for the assumptions of the statistical test. &~\pageref{reforms7c} & No formal p-value-based hypothesis tests were used; comparison is based on non-overlapping 95\%~CIs as per Agarwal et al.~(2021). \\
\midrule

\multicolumn{4}{l}{\textbf{Module 8: Generalizability and limitations}} \\
8a & Evidence of external validity. &  & Not reported. External validation was not performed as the study was conducted within simulated environments (UVA/Padova and LavaGap). \\
8b & Contexts in which the authors do not expect the study’s findings to hold. &~\pageref{reforms8b} & Results may vary in discrete clinical action spaces or high-dimensional real-world data where physiological features may be more robust to binning. \\
\midrule

\bottomrule
\end{longtable}

\end{document}